%% file: main.tex
\documentclass[10pt,journal,compsoc]{IEEEtran}
\newif\ifpeerreview

\peerreviewfalse

\usepackage{xcolor}
\usepackage[nocompress]{cite}
\usepackage{url}
\usepackage{amsmath,amssymb,graphicx}

\usepackage{lipsum} 
\usepackage{booktabs}
\usepackage[switch]{lineno}
\usepackage{multirow}
\newcommand{\paperID}{19}

\title{Ultrasound Tomography of Musculoskeletal Tissues with Generative Neural Physics}

\author{Zhijun~Zeng$^{*}$, Youjia~Zheng$^{*}$, Chang~Su$^{*}$, Qianhang~Wu, Hao~Hu, Zeyuan~Dong,
Yang~Lv, Ligang~Cui, Zhiyong~Hou, Weijun~Lin, Zuoqiang~Shi$^{\dagger}$,
Yubing~Li$^{\dagger}$,~\IEEEmembership{Member,~IEEE,} and He~Sun$^{\dagger}$,~\IEEEmembership{Member,~IEEE}%
\IEEEcompsocitemizethanks{
\IEEEcompsocthanksitem This work was supported in part by the National Key R\&D Program of China
(No.~2022YFC3401100), the National Natural Science Foundation of China
(Nos.~92370125, 12474461, 32450631, and 62371007), the Shanghai Municipal
Science and Technology Major Project (No.~2025SHZDZX026D03), the Joint
Research Project of the Shijiazhuang-Peking University Cooperation Program,
and the Basic and Frontier Exploration Project Independently Deployed by
Institute of Acoustics, Chinese Academy of Sciences (No.~JCQY202402).
\IEEEcompsocthanksitem $^{\dagger}$~Corresponding authors: He Sun; Yubing Li; Zuoqiang Shi.
E-mail: hesun@pku.edu.cn, liyubin@mail.ioa.ac.cn, zqshi@tsinghua.edu.cn.
\IEEEcompsocthanksitem $^{*}$~Zhijun Zeng, Youjia Zheng, and Chang Su contributed equally to this work.
\IEEEcompsocthanksitem Zhijun Zeng, Youjia Zheng, Qianhang Wu, Hao Hu, and He Sun
are with the College of Future Technology and the National Biomedical Imaging
Center, Peking University, Beijing 100871, China. Zhijun Zeng is also with the
Department of Mathematical Sciences, Tsinghua University, Beijing 100084, China.
\IEEEcompsocthanksitem Chang Su, Zeyuan Dong, Weijun Lin, and Yubing Li are with
the State Key Laboratory of Acoustics and Marine Information, Institute of
Acoustics, Chinese Academy of Sciences, Beijing 100190, China, and also with
the University of Chinese Academy of Sciences, Beijing 100049, China.
\IEEEcompsocthanksitem Yang Lv is with the Department of Orthopedics, Peking
University Third Hospital, Beijing 100191, China.
\IEEEcompsocthanksitem Ligang Cui is with the Department of Ultrasound, Peking
University Third Hospital, Beijing 100191, China.
\IEEEcompsocthanksitem Zhiyong Hou is with the Department of Orthopaedic Surgery,
The Third Hospital of Hebei Medical University, Shijiazhuang 050051, China.
\IEEEcompsocthanksitem Zuoqiang Shi is with the Yau Mathematical Sciences Center,
Tsinghua University, Beijing 100084, China, and also with the Yanqi Lake Beijing
Institute of Mathematical Sciences and Applications, Beijing 101408, China.}%
}

\begin{document}

\IEEEtitleabstractindextext{%
\begin{abstract}
Ultrasound Tomography (UT) is a radiation‐free, high‐resolution modality, but remains limited for musculoskeletal imaging due to the high computational cost and instability of full-waveform inversion in strongly scattering media. We propose a generative neural physics framework that couples generative networks with physics‐informed neural simulation for fast, high‐fidelity 3D UT. By learning a compact surrogate of ultrasonic wave propagation from a limited set of cross-modality images, our method merges the accuracy of wave modeling with the efficiency and stability of deep learning. This enables accurate quantitative imaging of \textit{in vivo} musculoskeletal tissues, producing spatial maps of acoustic properties beyond reflection‐mode images. 
On synthetic and \textit{in vivo} data of breasts, arms, and legs, we reconstruct 3D maps of tissue parameters in under ten minutes, with sensitivity to acoustic variations in musculoskeletal tissues and resolution comparable to MRI.  By overcoming computational bottlenecks in strongly scattering regimes, this approach demonstrates the feasibility of quantitative UT for musculoskeletal imaging and advances its development toward future routine clinical use.
\end{abstract}

}

\maketitle

\IEEEraisesectionheading{
  \section{Introduction}\label{sec:introduction}
}

\begin{figure*}[!htbp]
\centering
\framebox[\textwidth]{\parbox{0.95\textwidth}{\includegraphics[width=0.95\textwidth]{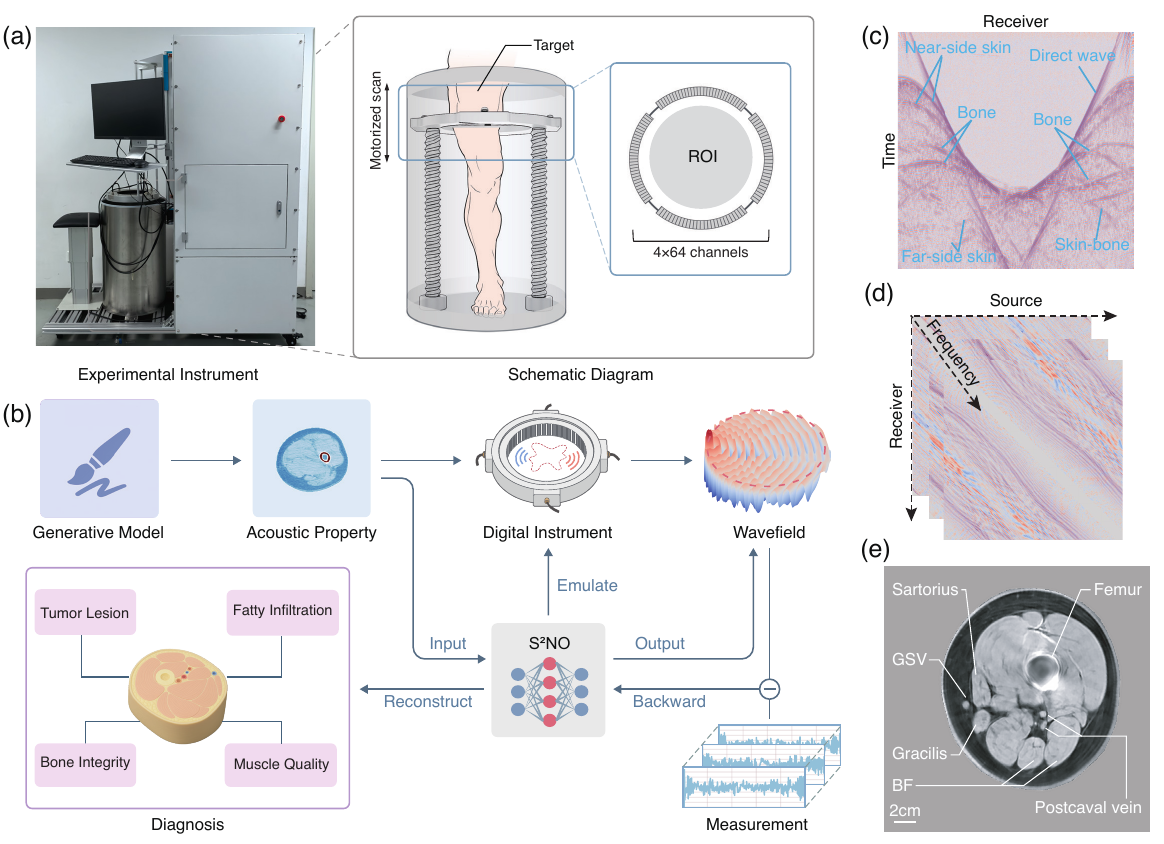}}}
\caption{(a) Illustration of the experimental setup for ultrasound tomography.
(b) Schematic overview of the proposed generative neural physics framework.
(c) Time-domain amplitude data of multiple receiver channels and single source acquired from an \textit{in vivo} human leg experiment.
(d) Multiple discrete frequency-domain datasets obtained after preprocessing. The color represents the real part of the complex amplitude.
(e) Transverse slices from the reconstruction of a female leg produced by the proposed framework. BF, biceps femoris; GSV, great saphenous vein.}
\label{fig:fig1}
\end{figure*}

\IEEEPARstart{U}{ltrasound} Tomography (UT) is a promising medical imaging technology that combines the safety, portability, and affordability of traditional ultrasound with high-resolution 3D imaging capabilities. Unlike conventional B-mode ultrasound\cite{rumack2017diagnostic}, which forms images solely based on echoes from tissue interfaces, UT utilizes specialized transducer arrays—often annular, cylindrical, or hemispherical—to collect both transmitted and reflected waves from the scattering media (Fig.~\ref{fig:fig1}). The UT array emits waves sequentially from each transducer and records the corresponding signals from all transducers, enabling it to reconstruct the internal structure of biological tissues in a tomographic fashion\cite{cueto2021spatial}. This technique is particularly advantageous for imaging tissues with high contrast and high resolution and for detecting abnormalities such as tumors\cite{ali20242,li2009vivo,ozmen2015comparing}. Moreover, UT avoids ionizing radiation and magnetic fields, making it safer for vulnerable populations\cite{guasch2020full}.

    Despite these advantages, the clinical adoption of UT has been limited by the immense computational demands associated with its image reconstruction process. Decoding the measured wavefield data into acoustic properties naturally leads to full-waveform inversion (FWI)\cite{tarantola1984inversion}, a nonlinear Partial Differential Equation (PDE)-constrained optimization problem that estimates spatial maps of acoustic parameters by minimizing the misfit between recorded and simulated wavefields, with spatial resolution targeted to approach subwavelength scales:
\begin{equation}\label{eq:fwi}
\begin{array}{cl}
  \min_{c, u_k}   &  \sum_{k=1}^K \| y_k - u_k(x_f) \|_2^2  \\
  \text{s.t.}   & \left[ \nabla^2 + \left( \frac{\omega}{c(x)} \right)^2 \right] u_k(x) = -\rho_k(x)
\end{array},
\end{equation}
where \( c \) is the sound speed distribution in tissue, \( u_k \) the 
acoustic wavefield, \( y_k \in \mathbb{C}^K \) the recorded data from $K$ transducers, \( x \) the spatial coordinates in the computational domain, \( x_f \) the transducer locations, \( \omega \)  the angular frequency, \( \rho_k \) the source terms, and \( k \) the transducer index. FWI requires repeatedly solving wave equations in both forward and inverse steps to iteratively update the medium model until simulated and measured data agree. Solving this problem is computationally demanding, especially in bone-containing tissues, where the high acoustic contrast between bone and surrounding soft tissues (e.g., muscle and fat) induces strong organ-scale multiple scattering. Under these conditions, current numerical methods remain inadequate for practical applications. Capturing high-frequency wave phenomena further requires prohibitively fine spatial resolution, leading to excessive computational cost and degraded numerical stability\cite{ernst2011difficult}. These limitations effectively preclude clinical UT in many settings, especially for large-scale 3D applications such as full-leg imaging for musculoskeletal disorders or whole-brain imaging for stroke assessment\cite{guasch2020full}. Although recent studies have demonstrated that chemical interventions, such as skull decalcification, can attenuate scattering in bone-containing tissues, these techniques remain confined to animal experiments, and their safety and applicability in human imaging have yet to be rigorously evaluated \cite{wang2025acoustic}.

 Here we propose a generative neural physics framework that integrates generative AI with neural PDE solvers to enable rapid, high-fidelity 3D UT imaging (Fig.~\ref{fig:fig1}b). The framework addresses two longstanding challenges in quantitative biomechanical imaging with deep learning: accurate, efficient modeling of organ-scale strong scattering, and the simulation-to-reality gap between training data and clinical use. However, existing methods remain inadequate in addressing these challenges in a unified manner. In particular, deep PDE solvers \cite{karniadakis2021physics,raissi2019physics} enforce governing equations by embedding PDE constraints into the training loss. However, they often fail to accurately capture highly oscillatory wave phenomena due to the inherent low-frequency bias of neural networks\cite{wang2022and}. By contrast, end-to-end inversion networks \cite{fan2022model,lozenski2024learned,molinaro2023neural,ren2024deep} learn direct mappings from measured wavefields to tissue properties without explicit PDE solvers, yet frequently overfit to simplified scattering media or restricted domains, yielding artifacts that undermine clinical reliability\cite{jin2024empirical,mcgreivy2024weak}.

 Our approach begins by generating anatomically realistic, large-scale human organ phantoms with accurate acoustic properties using physics-based style transfer of CT images and generative data augmentation. These phantoms drive a high-fidelity UT simulator to produce multi-frequency wavefields. We then introduce the Strong Scattering Neural Operator ($S^{2}NO$), designed for highly oscillatory PDEs on large domains. By embedding wave-propagation physics into its architecture, $S^{2}NO$ accurately models multiple scattering in clinical settings, achieving orders-of-magnitude speedups over conventional solvers and establishing itself as a versatile wave foundation model. Integrating $S^{2}NO$ into a neural FWI framework enables \textit{in vivo} high-resolution 3D UT of bone-containing human tissues—long considered beyond the reach of ultrasound—and delivers MRI-comparable image quality (Fig.~\ref{fig:fig1}e). This defines a Reality-to-Simulation-to-Reality (Real2Sim2Real) pipeline: (1) constructing realistic UT simulation databases from limited dataset of patient CT scans (Real2Sim); (2) achieving \textit{in vivo} high-resolution UT imaging using a physics-informed neural PDE solver trained exclusively on synthetic data (Sim2Real). Moreover, we reduce the computational time from several hours to under 10 minutes without compromising resolution, thereby paving the way for the broad clinical adoption of UT.
\begin{figure*}[!t]
\centering
\framebox[\textwidth]{\parbox{0.95\textwidth}{\includegraphics[width=0.95\textwidth]{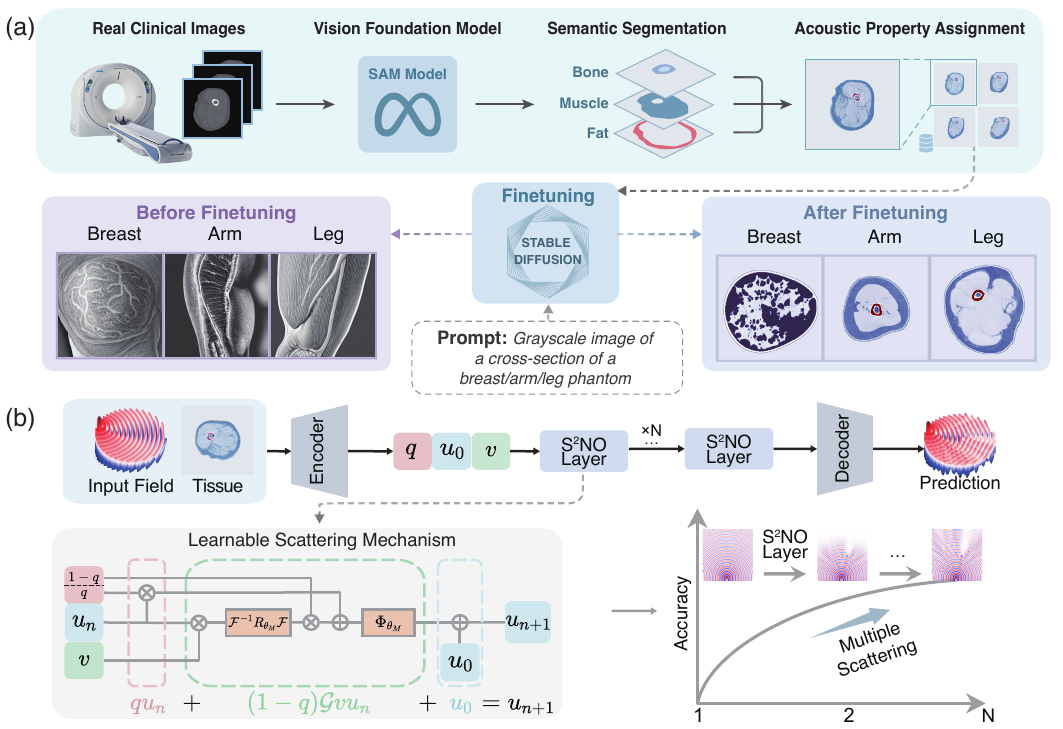}}}
\caption{(a) Schematic overview of the UT dataset generation pipeline. (b) Schematic overview of the entire $S^2NO$ architecture.}
\label{fig:fig2}
\end{figure*}

\section{Method}\label{sec:method}
In this section, we present a generative neural physics framework that couples generative AI with a physics-informed PDE solver for fast, high-fidelity UT.
The framework integrates a generative model–based Real2Sim data pipeline, a strong scattering neural operator ($S^2$NO) derived from the convergent Born series, and an $S^2$NO-based FWI scheme that alleviates the computational cost of PDE solves for efficient UT reconstruction.

\subsection{Generation of UT Dataset with Realistic Anatomy and Experimental Scenario}
The success of deep learning models for medical imaging hinges on access to large amounts of realistic training data. Unlike CT or MRI, conventional ultrasound lacks standardized tomographic anatomy image databases, and although UT offers a promising solution, limited system deployment and inefficient solvers make large-scale human organ dataset acquisition impractical. To address this challenge, we developed an innovative Real2Sim strategy that combines cross-modality clinical images with a foundation generative model to produce anatomically and physically realistic UT phantoms (Fig.~\ref{fig:fig2}a) and corresponding wavefields simulated using parameters from an actual UT system (Fig.~\ref{fig:fig1}a). Our dataset comprises 22,047 digital human organ phantoms, including 7,520 breasts, 7,526 arms and 7,001 legs. For each phantom, we simulate wavefields from 64 different sources at 8 different frequencies, resulting in a total of 11,288,064 input-output data pairs. Our dataset is released at \url{https://ai4scientificimaging.org/OpenWaves}

\subsubsection{Physics-based Generation of Organ Phantoms}
For breast phantoms, we employ the Virtual Imaging Clinical Trial for Regulatory Evaluation (VICTRE) framework developed by the U.S. Food and Drug Administration (FDA) to generate digital 3D anatomical models \cite{badano2018evaluation}. Following \cite{li20213}, tissue-dependent sound speed distributions are assigned according to breast density categories, including all-fatty (FAT), fibroglandular (FIB), heterogeneous (HET), and extremely dense (EXD) types. The generation pipeline consists of three steps:
\begin{enumerate}
    \item 3D breast anatomies are synthesized using the VICTRE tool.
    \item 2D phantoms are extracted via random cross-sectional slicing, followed by spatial normalization and random scaling to increase diversity.
    \item Distinct breast tissues (e.g., skin, adipose tissue, and muscle) are segmented and physically realistic sound speeds, each with small random perturbations, are assigned to the corresponding tissue regions. The region surrounding the breast models is then filled with water to replicate experimental conditions. We applied additional random rotations to the samples to enhance the diversity of the data.
\end{enumerate}
This process yields 3,520 anatomically realistic breast phantoms. However, the high computational cost of VICTRE restricts the range of anatomical variations that can be explored, limiting dataset diversity.

\begin{figure*}[!htbp]
\centering
\framebox[\textwidth]{\parbox{0.95\textwidth}{\includegraphics[width=0.95\textwidth]{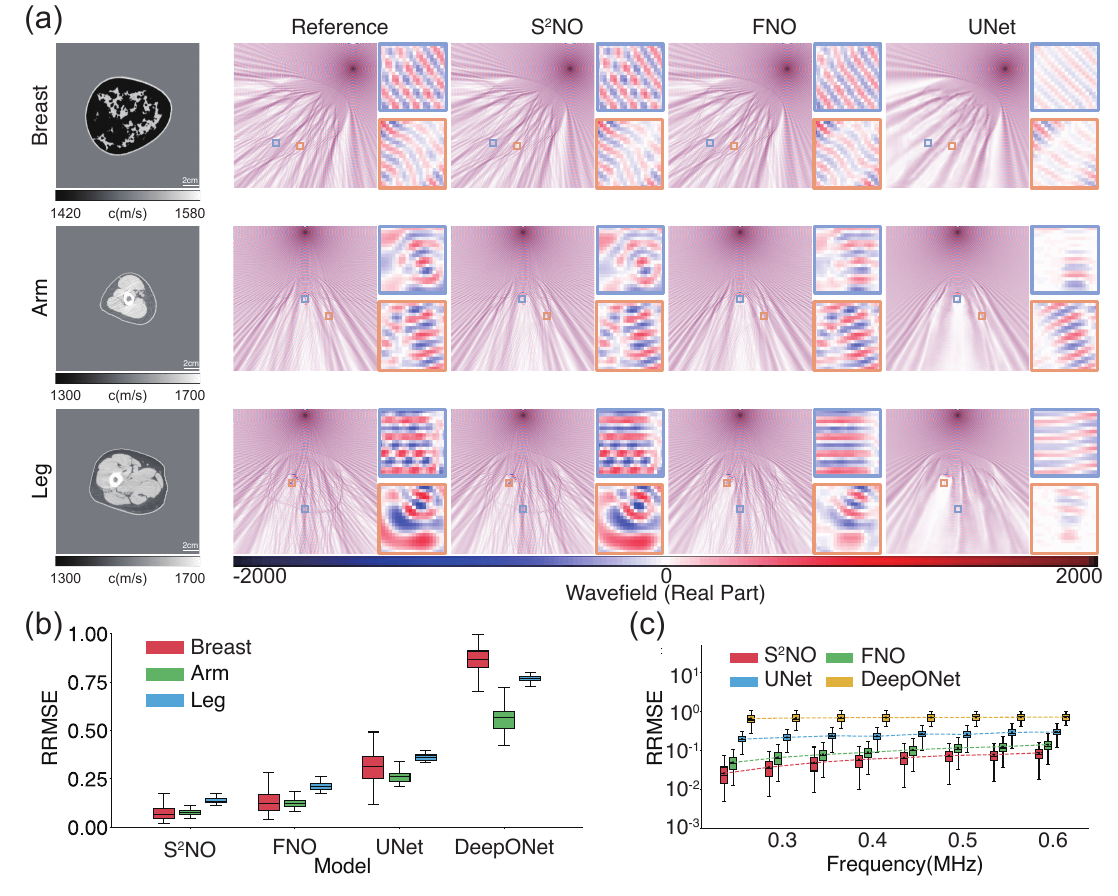}}}
\caption{(a) Wavefield predictions from $S^2NO$ and baseline models for breast, arm, and leg phantoms at 0.6MHz. (b) Forward simulation errors of $S^2NO$ and baseline models for breast, arm, and leg phantoms at a 0.6MHz setting. Boxplots show the median (central line) and interquartile range (IQR, box); whiskers extend to the smallest and largest values within 1.5× the IQR (outliers omitted), sample size N=320/373/108 for Breast/Arm/Leg slices dataset. (c) Comparison of forward simulation errors between $S^2NO$ and the baseline models for all organ phantoms at frequencies of 0.25$\sim$0.6 MHz. The
setting of the boxplots is the same as (b), sample size N=801.}
\label{fig:fig3}
\end{figure*}

For arm and leg phantoms, we develop a physics-inspired modality transfer pipeline from CT images due to the absence of dedicated simulation tools for musculoskeletal structures.  The CT data were acquired from consenting volunteers recruited from the Orthopedic Department of Peking University Third Hospital, including 151 arm scans and 41 leg scans. Informed consent was obtained from all participants prior to data collection. The study was approved by the Peking University Third Hospital Medical Science Research Ethics Committee (Project M2022262) and was conducted in accordance with all relevant ethical standards.
The selected slices are converted into sound speed maps through the following procedure:
\begin{enumerate}
    \item CT images are semantically segmented using the SAM foundation model \cite{kirillov2023segment}.
    \item Each anatomical region is mapped to a physically plausible sound speed range.
    \item The anatomy is embedded into a water background, and random geometric transformations are applied to increase diversity.
\end{enumerate}
This pipeline produces 809 arm and 1,001 leg phantoms. Due to structural similarity across slices from the same subject, the resulting dataset exhibits limited diversity.


\subsubsection{Fine-tuning Stable Diffusion for Data Augmentation}

To address the limited diversity of physics-based phantoms, we introduce a generative augmentation strategy by fine-tuning a pretrained text-to-image diffusion model.

For each organ type, a Stable Diffusion model is fine-tuned using the DreamBooth method \cite{ruiz2023dreambooth} on a subset of 2D phantom slices, with the prompt “Grayscale image of a cross-section of a breast/arm/leg phantom.” This enables the model to capture organ-specific anatomical distributions beyond the limited physics-based samples. The augmentation pipeline consists of three steps:
\begin{enumerate}
    \item The physics-based phantoms (breast, arm, and leg) are used to fine-tune the pretrained model, enabling adaptation to the target anatomical domains.
    \item Generated samples are filtered based on physical plausibility criteria. Specifically, we enforce boundary integrity and smoothness to ensure closed and non-fragmented anatomical structures, constrain anatomical topology to preserve realistic bone geometry and muscle organization, and remove samples exhibiting inconsistent or physically implausible tissue configurations. Details of the data filtering procedure are provided in the Supplementary Material.
    \item The retained samples are aligned with the physics-based simulation pipeline. This includes normalization to a common field-of-view (FOV), alignment of anatomical structures, and assignment of sound speed values using the same tissue-dependent rules as the physics-based phantoms.
\end{enumerate}
Using this procedure, we generate 4,000 breast, 6,717 arm, and 6,000 leg phantoms. As shown in Fig.~\ref{fig:fig2}a, the fine-tuned model produces anatomically consistent structures, whereas the pretrained model often yields semantically plausible but physically unrealistic outputs.

\subsubsection{Wavefield Simulation using Experimental UT Settings}
Wavefields in our dataset were simulated using the experimental parameters of two actual annular UT systems. Specifically, the breast dataset employed the system configuration described by Ali et al\cite{ali20242}, whereas the arm and leg datasets utilized a system of similar size but with a distinct transducer arrangement. Both UT systems comprise 256 transducers uniformly distributed around a ring with a diameter of around 22 centimeters. The system operates within a frequency range of 0.25 MHz to 1.2 MHz, corresponding to approximate acoustic wavelengths of 1 mm to 5 mm. The use of sub-MHz frequencies is motivated by waveform-based transmission UT, where the frequency must balance resolution, penetration depth, and inversion stability. Unlike conventional reflection-mode clinical ultrasound systems operating at 3--5 MHz or higher, transmission UT relies on coherent wave propagation across the imaging domain. For the target resolution of approximately 1 mm considered in this study, sub-MHz data already provide an appropriate half-wavelength scale for speed-of-sound reconstruction. Higher-frequency measurements, while potentially improving nominal resolution, would undergo stronger attenuation and scattering and introduce greater phase ambiguity, thereby making FWI more prone to cycle skipping and unstable speed estimation.
In our simulations, we focus on wave propagation at discrete frequencies of 0.25, 0.3, 0.35, 0.4, 0.45, 0.5, 0.55, and 0.6 MHz, yielding approximately 40 to 100 wavenumbers within the field of view. Consistent with standard frequency-domain FWI, the inversion is performed by updating the model at these discrete frequency points \cite{pratt1999seismic,ali20242} with a frequency-marching strategy that sequentially optimizes from low to high frequencies. For each phantom, we simulated 64 wavefields corresponding to distinct point sources, uniformly selected from the 256 transducers by employing every fourth transducer as the source location. The Helmholtz equations are solved numerically using the Convergent Born Series\cite{osnabrugge2016convergent} (CBS) algorithm. Together, the synthetic phantoms and simulated measurements provide a critical foundation for training and evaluating next-generation UT imaging algorithms.

\subsection{Strong Scattering Neural Operator ($S^2NO$)}
The $S^2NO$ is designed to accurately and efficiently approximate the implicit mapping $f_{c,\rho}:c,\rho\rightarrow u$ as defined by the Helmholtz equation. Drawing inspiration from the CBS solver for the Helmholtz equation, the $S^2NO$ utilizes an iterative scheme in developing its neural network architecture. Moreover, the inputs and outputs within each iterative block of the $S^2NO$ are carefully designed based on physical concepts, thereby enhancing the operator's ability to approximate and generalize effectively.
\subsubsection{Born Series and Convergent Born Series}
By defining a scattering potential
$
v=\left(\frac{\omega}{c(x)}\right)^2-\kappa^2-i\varepsilon,
$
 where $\kappa$ denotes a reference (background) wavenumber and $\varepsilon$ is a small positive damping parameter introduced to ensure convergence, we can reformulate the Helmholtz equation as
\begin{equation}
\mathcal{S}'u(x)\triangleq\big[\nabla^{2}+\kappa^{2}+i\varepsilon\big]\,u(x)
= -\rho(x)-v(x)\,u(x).
\end{equation}
This leads to the derivation of the Born series for the Helmholtz equation:
\begin{equation}
u=\mathcal{G}\rho+\mathcal{G} v\,u
\;\Rightarrow\;
u=\sum_{n=0}^{\infty}(\mathcal{G}v)^{n}\,\mathcal{G}\rho .
\end{equation}
where $\mathcal{G}=\mathcal{F}^{-1}\!\circ\!(p^{2}-\kappa^{2}-i\varepsilon)^{-1}\!\circ\!\mathcal{F}$ is the Green operator of
$\mathcal{S}'(\cdot)=\big[\nabla^{2}+\kappa^{2}+i\varepsilon\big](\cdot)$, $\mathcal{F}$ is the Fourier transform operator, and $p^{2}=p_x^{2}+p_y^{2}$ is the squared magnitude of the Fourier-transformed coordinates. This series solution can also be used to derive an iterative form of the solution
\begin{equation}
\left\{\begin{array}{ccl}
u_{n+1}&=&u_{0}+(\mathcal{G}v)\,u_{n},\\
u_{0}&=&\mathcal{G}\rho .
\end{array}\right.
\end{equation}
While the Born series has been effective for solving Helmholtz equations with weak scattering potentials, its convergence is limited in the presence of strongly scattering samples. This poses challenges for practical wave simulations, including biomedical UT.

To address this limitation, Osnabrugge et al.\cite{osnabrugge2016convergent} proposed a modification to the Born series by introducing a preconditioner
$q=1-\frac{i\,v}{\varepsilon}$. This approach involves multiplying both sides with $1-q$ and introducing a new operator
$\mathcal{M}=(1-q)\,\mathcal{G}v+q$, leading to a new formulation
\begin{equation}
\begin{aligned}
(1-q)u &= (1-q)\,\mathcal{G}\rho + (1-q)\,\mathcal{G}v u, \\
\Rightarrow u &= \sum_{n=0}^{\infty} \mathcal{M}^{n}\big((1-q)\,\mathcal{G}\rho\big).
\end{aligned}
\end{equation}
which yields the iterative scheme
\begin{equation}
\left\{\begin{array}{ccl}
u_{n+1}&=&u_{0}+\mathcal{M}\,u_{n},\\
u_{0}&=&(1-q)\,\mathcal{G}\rho .
\end{array}\right.
\end{equation}
Here, the operator $\mathcal{M}$ maintains a spectral radius smaller than $1$ with a suitable choice of $\varepsilon$ and $\kappa$, ensuring convergence for strong scattering potentials.

\begin{figure*}[!t]
\centering
\framebox[\textwidth]{\parbox{0.95\textwidth}{\includegraphics[width=0.95\textwidth]{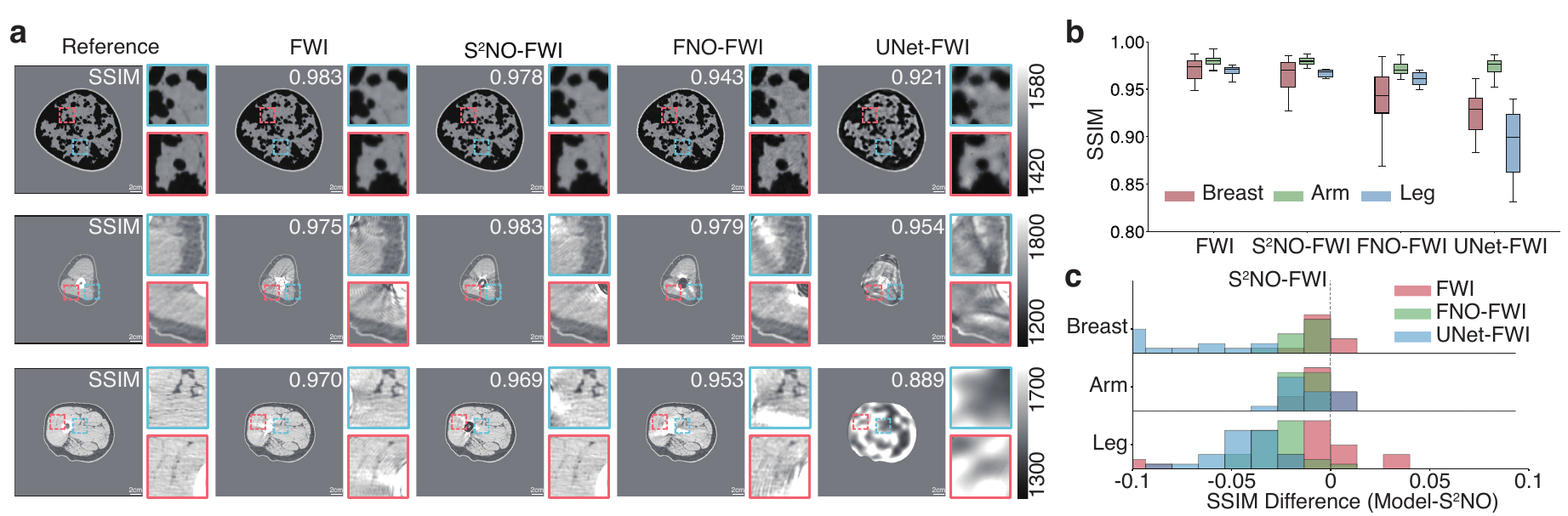}}}
\caption{(a) Comparison between the reference synthetic breast, arm and leg phantoms and the reconstructed phantoms using different models (numerical solver, $S^2NO$, FNO and UNet). (b) Statistical summary of SSIM values for reconstruction results of breast, arm, and leg phantom in test dataset using various models (numerical solver, $S^2NO$, FNO and UNet). Boxplots show the median (central line) and interquartile range (IQR, box); whiskers extend to the smallest and largest values within 1.5× the IQR (outliers omitted), sample size N=24/16/12 for Breast/Arm/Leg dataset.
(c) Distribution of SSIM differences between $S^2NO$ and other methods computed on a sample-by-sample basis.}
\label{fig:fig4}
\end{figure*}

\subsubsection{$S^2NO$ Architecture}
The $S^2NO$’s network architecture is crafted based on the iterative structure
of the CBS. As depicted in Fig.~\ref{fig:fig2}b, $S^2NO$ comprises three distinct phases:
1) encoding the sound speed $c$ and the source $\rho$ into a latent space to
obtain the representations of $q$, $v$ and $u_{0}$;
2) iteratively updating the latent wavefield states $u_n$ using $S^2NO$ layers; and
3) projecting the final latent state back into the physical space.
This architecture is mathematically represented as follows:
\begin{equation}
\begin{aligned}
\text{Phase 1:}\quad & q=\mathcal{E}_{\theta_q}(c),\quad
v=\mathcal{E}_{\theta_v}(c),\quad
u_{0}=\mathcal{E}_{\theta_{\text{init}}}(c,\rho),\\
\text{Phase 2:}\quad & u_{n+1}=\mathcal{M}_{\theta_M}(u_n,q,v)+u_0,\quad n=0,\ldots,N,\\
\text{Phase 3:}\quad & u_{\text{out}}=\mathcal{D}_{\theta_{\text{pred}}}(u_N).
\end{aligned}
\end{equation}
In this model, $q$ and $v$ are the latent states of the preconditioner and the scattering potential, $u_n$ is the latent state of the wavefield, and $u_{\text{out}}$ is the output wavefield in the physical space. The encoders $\mathcal{E}_{\theta_q}$, $\mathcal{E}_{\theta_v}$ and $\mathcal{E}_{\theta_{\text{init}}}$ process the auxiliary variables of the preconditioner, the scattering potential and the initial wavefield, respectively, while $\mathcal{M}_{\theta_M}$ acts as the operator $\mathcal{M}$ to update the latent wavefield. Indeed, we replace the Green operator $\mathcal{G}$ in $\mathcal{M}$ with a spectral convolution layer, obtaining the $S^2NO$ layer
$\mathcal{M}_{\theta_M}$:
\begin{equation}
\mathcal{M}_{\theta_M}(u_n,q,v)=
\Phi_{\theta_M}\!\left((1-q)\,\mathcal{F}^{-1}\!\circ R_{\theta_M}\!\circ
\mathcal{F}(u_n v)+q\,u_n\right).
\end{equation}
Here, $\Phi_{\theta_M}$ is a CNN module and $R_{\theta_M}$ represents the trainable Fourier modes\cite{lifourier}. Each $S^2NO$ layer uses independent parameters to boost the network’s representational capacity. The decoder $\mathcal{D}_{\theta_{\text{pred}}}$ then predicts the wavefield.
\subsection{$S^2NO$-based FWI}
The PDE-constrained optimization problem in \eqref{eq:fwi} is typically solved using gradient-based methods with adjoint-based gradient computation. Using the method of Lagrange multipliers, the constrained optimization problem can be reformulated as an unconstrained form:
\begin{equation}
\begin{array}{cll}
    \min\limits_{c,\,u_k,\,\lambda}\ \mathcal{L}&
=& \sum_{k=1}^{K} \mathcal{L}_k
= \sum_{k=1}^{K}\bigl\|y_k-u_k(x_f)\bigr\|_2^2\\
&-& \sum_{k=1}^{K}\Big\langle \lambda_k(x),\,\bigl[\nabla^2+\bigl(\tfrac{\omega}{c(x)}\bigr)^2\bigr]\,u_k(x)+\rho_k(x) \Big\rangle .
\end{array}
\end{equation}
where $\mathcal{L}$ is the Lagrangian function, $\langle f,g\rangle$ denotes the
real part of the inner product of functions $f$ and $g$ in $L^2(\mathbb{C})$.
The gradient $\frac{\partial \mathcal{L}_k}{\partial c}(x)$ is proportional to
the product of $u_k$ and $\lambda_k$:
\begin{equation}
\frac{\partial \mathcal{L}_k}{\partial c}(x)
= -2\omega^{2}\operatorname{Re}\!\left(\frac{\lambda_k^{*}(x)\,u_k(x)}{c(x)^3}\right),
\end{equation}
where the wavefield $u_k$ and the adjoint state $\lambda_k$ are obtained by solving
the Helmholtz equation
\begin{equation}\label{eq:adjoint}
\left\{
\begin{aligned}
\bigl[\nabla^2+\bigl(\tfrac{\omega}{c(x)}\bigr)^2\bigr]\,u_k(x) &= -\rho_k(x),\\
\bigl[\nabla^2+\bigl(\tfrac{\omega}{c(x)}\bigr)^2\bigr]\,\lambda_k(x)
&= 2\sum_{i=1}^{K}\Big[u_k\!\bigl(x_f^{(i)}\bigr)-y_k^{(i)}\Big]\,
\delta\!\bigl(x_f^{(i)}\bigr).
\end{aligned}
\right.
\end{equation}
By incorporating $S^2NO$ as an approximation of the solution maps
$f_{c,\rho}(c(x),\rho(x))\to u(x)$, \eqref{eq:adjoint} can be approximated by
\begin{equation}
\left\{
\begin{aligned}
u_k(x) &= S^2NO\bigl(c(x),\,\rho_k(x)\bigr),\\
\lambda_k(x) &= -2\sum_{i=1}^{K}S^2NO\!\left(c(x),\,\Big[u_k\!\bigl(x_f^{(i)}\bigr)-y_k^{(i)}\Big]\,
\delta\!\bigl(x_f^{(i)}\bigr)\right).
\end{aligned}
\right.
\end{equation}
Finally, the optimization problem is solved iteratively using gradient-based
optimization methods (e.g., L-BFGS, NCG).

\section{Experimental Results}
In this section, we evaluate the proposed generative neural physics framework for UT on both synthetic and \textit{in vivo} datasets. Experimental acquisition protocols for UT and MRI data are first introduced, followed by the training setup for the neural operator models. We then assess the performance of $S^{2}NO$ in forward wavefield simulation and its effectiveness in inverse UT image reconstruction, with comparisons to conventional numerical solvers and neural operator baselines. Finally, the proposed framework is validated on \textit{in vivo} data of multiple human organs, including breast, arm, and leg, demonstrating its ability to achieve accurate and efficient 3D imaging of \textit{in vivo} human tissues.
\subsection{Experimental Setting}\label{subsec:exp}
\subsubsection{Experimental Data Acquisition}

\textit{In vivo} breast UT data were obtained from the Karmanos Cancer Institute (IRB Approval No. 040912M1) \cite{ali20242}, including cases with benign and malignant lesions. \textit{In vivo} UT data acquisition for human arms and legs was performed using a circular transducer array (Fig.~\ref{fig:fig1}a). Axial scanning was achieved by vertically translating the system using retractable bellows, enabling layer-by-layer measurements. In this annular-array UT system, each transducer element acts as a small planar source rather than an ideal point source, resulting in a directivity effect that largely confines wave propagation within a 2D slice. Consequently, 2D modeling provides an effective approximation of the imaging process. The system consists of 256 transducers uniformly arranged in a circular geometry surrounding the region of interest. During acquisition, one transducer emits an acoustic signal while all elements record the responses. This procedure is repeated sequentially for all transducers, yielding a complete set of measurements for each slice. By scanning the arm or leg along the axial direction, multiple 2D slices are acquired and subsequently stacked to form a 3D volume. \textit{In vivo} UT experiments on the arm and leg were conducted on two adult volunteers, one male and one female, at Peking University Third Hospital. The data were collected with informed consent from all participants under the approval of the Institutional Review Board of Peking University Third Hospital (IRB Approval No. IRB00006761-M2024690), and all procedures were performed in accordance with relevant ethical regulations.

For anatomical reference, MRI data were acquired using a GE Discovery MR750W 3.0T scanner equipped with an 8-channel PA Matrix coil. Participants were positioned supine, feet first, with the long axis of the lower limbs aligned with the scanner’s main axis. The coil was positioned to fully cover the patella for axial imaging. Scanning protocols included: axial fat-suppressed T2-weighted imaging (repetition time (TR) = 5169 ms and echo time (TE) = 72.1 ms) and axial T1-weighted imaging (TR = 618 ms, TE = 11.1 ms, FOV = 400 × 275.2 mm$^2$, slice thickness = 6 mm, inter-slice gap = 0.6 mm, matrix = 320 × 256). 
\begin{figure*}[htbp]
\centering
\framebox[\textwidth]{\parbox{0.95\textwidth}{\includegraphics[width=0.95\textwidth]{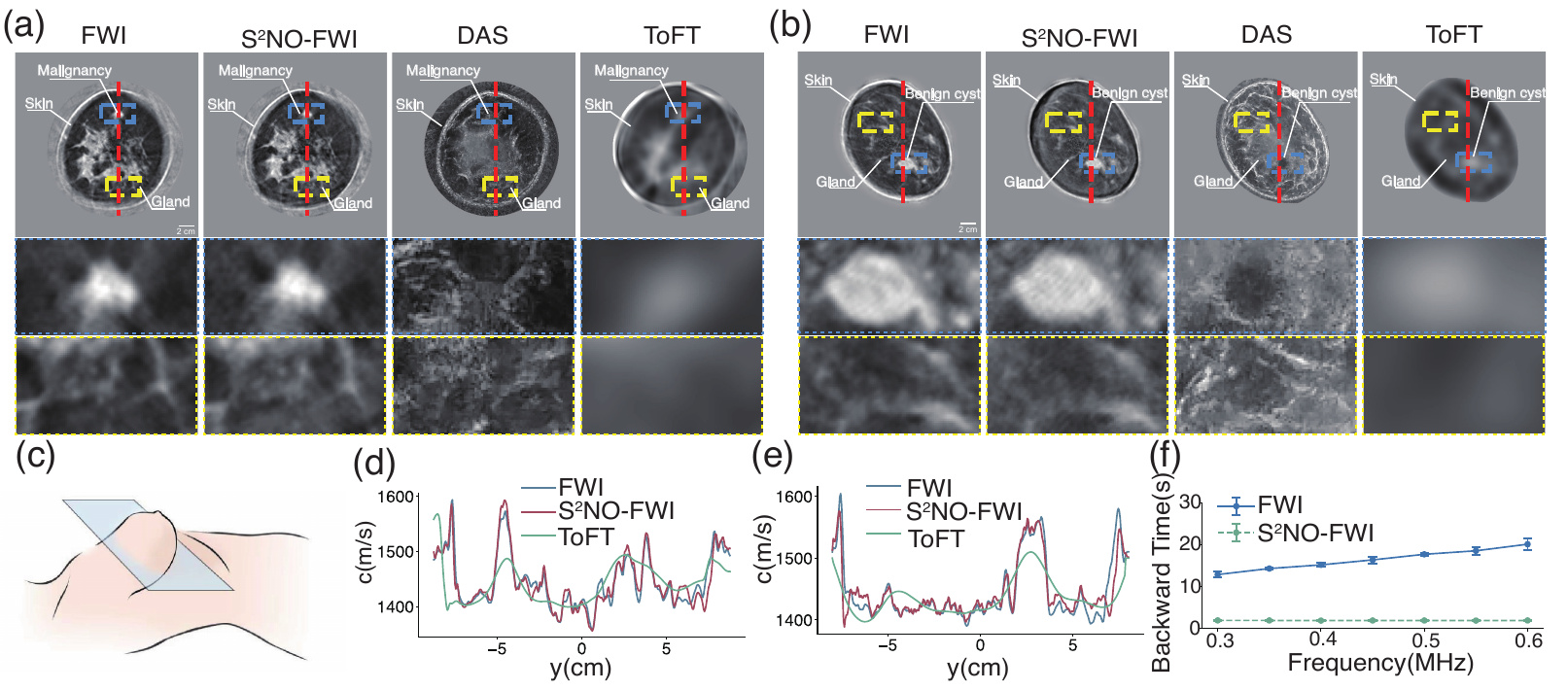}}}
\caption{(a)-(b) Comparison of reconstructions of breasts with a spiculated malignancy (a) and a benign cyst (b), obtained using various imaging techniques (FWI, $S^2NO$-FWI, ToFT, and DAS). (c) Schematic illustration of cross-sectional data acquisition for breast imaging.
(d)-(e) One-dimensional velocity profiles extracted at specific positions (red line) from the reconstructed breast model with a spiculated malignancy (d) and a benign cyst (e).
(f) Computational time per iteration for traditional solver-based FWI and $S^2NO$-FWI across varying frequencies. The data points denote mean values, and the vertical bars indicate the standard deviation, sample size N=10.}
\label{fig:fig5}
\end{figure*}

\subsubsection{Neural Network Training}
For diffusion-based data augmentation, we fine-tuned a pretrained Stable Diffusion v1.4 model using the DreamBooth method \cite{ruiz2023dreambooth}. Specifically, 2D phantom slices of the breast, arm, and leg generated by the physics-based pipeline were converted to grayscale images and used for training, with the breast data balanced across the four density categories. Training hyperparameters are provided in the Supplementary Material. During inference, image synthesis was performed using the DDIM sampler with the same text prompt as in training.

For wavefield simulation, we compare the proposed $S^2NO$ with three baseline models: Fourier Neural Operator (FNO) \cite{lifourier}, DeepONet \cite{lu2021learning}, and U-Net \cite{ronneberger2015u}. All models were trained and evaluated under a unified protocol.  For all methods,  we trained an independent network for each discrete frequency. This frequency-specific training strategy can be viewed as a lightweight mixture-of-experts design, where each network specializes in a particular wavelength regime.  For each frequency, the training set consisted of wavefields generated from 64 uniformly sampled source locations selected from the 256 available transducers (consistent with the corresponding experimental acquisition setup). All models were trained on the combined dataset of breast, arm, and leg phantoms for 30 epochs using the AdamW optimizer, with an initial learning rate of $5\times10^{-3}$ and a StepLR decay schedule. Performance was evaluated using the relative root-mean-square error (RRMSE),
\begin{equation}
\mathrm{RRMSE} = \frac{1}{M}\sum_{i=1}^{M} \frac{\|u_i - \hat{u}_i\|_{2}}{\|u_i\|_{2}}.
\end{equation}
All experiments were conducted on four NVIDIA A800 GPUs (80 GB each). Additional implementation details are provided in the Supplementary Material.
\begin{figure*}[t]
\centering
\framebox[\textwidth]{\parbox{0.9\textwidth}{\includegraphics[width=0.9\textwidth]{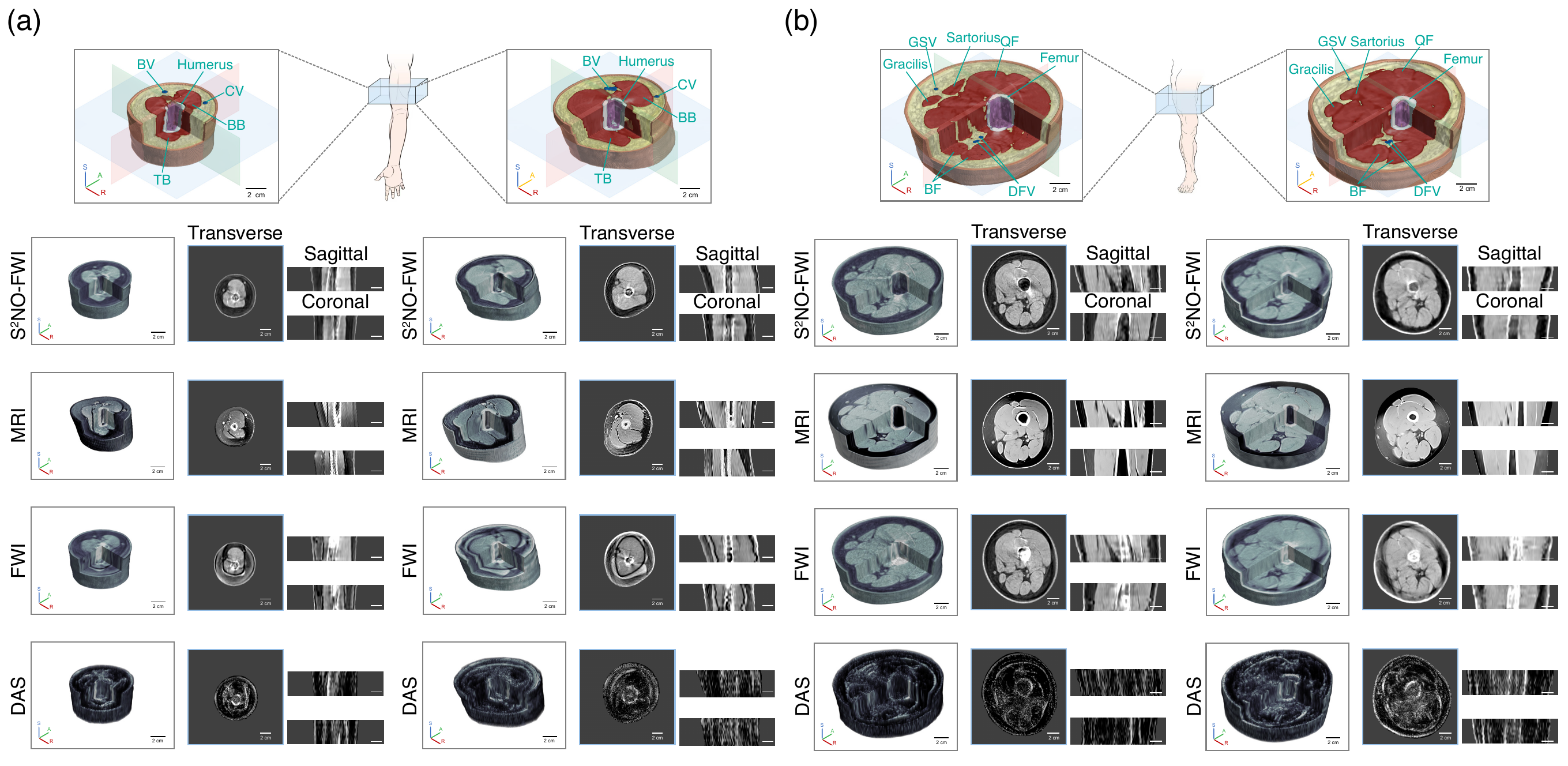}}}
\caption{Three-dimensional UT imaging of the arms and legs of a female subject (a–b, left) and a male subject (a–b, right).
The top row presents 3D UT reconstructions of the arms and legs with color-coded anatomical segmentations produced by the proposed $S^2NO$-FWI method.
The bottom row shows 3D reconstructions of the arms (a) and legs (b) obtained using $S^2$NO-FWI, MRI, conventional FWI, and delay-and-sum (DAS), assembled from 11 segmented slices. Representative transverse, sagittal, and coronal sections reconstructed by the different imaging methods are also provided.
BF, biceps femoris; QF, quadriceps femoris; DFV, deep femoral vein; GSV, great saphenous vein; BB, biceps brachii; TB, triceps brachii; BV, brachial vein; CV, cephalic vein.}
\label{fig:fig6}
\end{figure*}
\subsection{$S^2NO$ for Wave Simulation and UT Reconstruction}
 We first evaluated the accuracy of $S^2NO$ in simulating acoustic wavefields for synthetic breast, arm, and leg phantoms, using solutions from numerical solvers (CBS) as ground truth. Figure \ref{fig:fig3} compares wavefield predictions at 0.6 MHz for representative phantoms using various neural operators. $S^2NO$ outperformed the FNO, DeepONet and U-Net, accurately capturing complex scattering patterns, such as those at  the bone-muscle interface and within the marrow cavity of the femur.  While competing neural network methods struggled to handle highly oscillatory scenarios in PDE solutions, $S^2NO$ consistently achieved simulation accuracy comparable to traditional solvers. In addition to delivering superior accuracy, $S^2NO$ achieved significant computational speedups, 25$\sim$34 times faster than the CBS numerical solver for musculoskeletal tissues on an NVIDIA A800-80GB GPU, making it suitable for clinical UT trials. 

We then evaluated neural operator surrogates for image reconstruction from simulated observations of synthetic phantoms, as improved forward modeling accuracy is expected to yield more accurate inverse reconstructions. A gradient-based FWI approach derived from the adjoint method was employed. Ultrasonic signals at multiple frequencies (0.3 MHz to 0.6 MHz for breasts and 0.25 MHz to 0.6 MHz for arms and legs) were leveraged, progressively moving from low to high frequencies to enhance reconstruction accuracy\cite{bunks1995multiscale}. Further baseline comparisons at 500 kHz against recent neural operators, including U-NO~\cite{rahman2023uno}, BFNO~\cite{zhao2023deep}, GNOT~\cite{hao2023gnot}, ONO~\cite{xiao2024ono}, and FFNO~\cite{tran2023factorized}, as well as ablation studies on the architecture of $S^2NO$, are reported in the Supplementary Material, demonstrating the rationale and effectiveness of the proposed network design.

Figure \ref{fig:fig4} compares FWI reconstructions for breast, arm, and leg phantoms obtained using the conventional numerical solver (CBS), $S^2NO$, and other neural operators. Beyond a simple performance comparison, these simulated experiments serve to validate the numerical fidelity of $S^2NO$ as a surrogate forward solver for inverse reconstruction. The quantitative evaluation metric (SSIM) demonstrates that $S^2NO$-based FWI achieved reconstruction quality comparable to that of CBS and outperformed other neural operators across all phantom types (Fig.~\ref{fig:fig4}b,c). Specifically, $S^2NO$ effectively recovered anatomical features—including skin, muscle, fat, and bone—with enhanced clarity and delivered accurate quantitative acoustic contrasts. For breast phantoms exhibiting weaker scattering, $S^2NO$ predicted more accurate high-frequency information compared to other baseline methods, resulting in sharper resolution at interfaces. For arm and leg phantoms that exhibit stronger scattering, $S^2NO$ more accurately modeled multiple scattering phenomena, thus capturing bone morphology with improved fidelity and effectively suppressing scattering-induced imaging artifacts observed in other baselines’ reconstructions. Notably, during reconstruction, $S^2NO$ tended to underestimate bone thickness, whereas the conventional numerical solver more often overestimates it. Given its faster forward simulation, $S^2NO$-based FWI reduced the reconstruction time by up to fourteen-fold compared to traditional FWI methods.

\subsection{\textit{In Vivo} Human Breasts}
We apply $S^2NO$-based FWI to the \textit{in vivo} breast UT datasets described in Sec.~\ref{subsec:exp} to validate the model trained in Real2Sim stage. The datasets comprise breasts with benign and malignant lesions, from which acoustic signals at seven selected frequencies (0.3–0.6 MHz) are used for sound speed reconstruction.

As shown in Fig.~\ref{fig:fig5}, $S^2NO$-based reconstructions captured detailed breast structures, including skin, glandular tissue, tumors, ductal networks, and fatty regions. Compared to conventional delay-and-sum (DAS)\cite{karaman1995synthetic,thomenius1996evolution} and time-of-flight tomography (ToFT)\cite{duric2007detection} algorithms, $S^2NO$-based FWI produced higher-resolution images of interior structures with a parameter-free resolution\cite{descloux2019parameter} of approximately 1.36 mm, approaching the theoretical FWI limit (half the highest-frequency ultrasound wavelength) but cutting computation time 6-fold versus conventional numerical solvers. Importantly, Fig.~\ref{fig:fig5}d,e show that $S^2NO$-based FWI concretely distinguished malignant masses—characterized by irregular boundary morphology and a higher sound speed ($\sim$1590 m/s)—from benign lesions, which display regular boundaries and a lower sound speed ($\sim$1560 m/s). These observations highlight the capability of our method to recover fine-scale acoustic and morphological features in \textit{in vivo} breast UT.
Notably, the experimental apparatus used for imaging the breast with a benign cyst was not identical to the settings employed in the simulation data (e.g., transducer locations), revealing the exceptional generalization capability of $S^2NO$ to different hardware settings.
\begin{table}
    \centering
    \caption{\textbf{Runtime for inverse imaging of experimental data}}
    \label{tab:runtime_invivo_main}
\begin{tabular}{@{}l c c c@{}}
\hline
\multicolumn{4}{c}{Inverse Imaging of experimental data} \\
\hline
\multirow{2}{*}{Model}  & \multicolumn{3}{c}{Imaging process time [s]} \\
\cline{2-4}
& Breast & Arm & Leg \\
\hline
S$^{2}$NO  & $466.85\pm74.76$ & $510.29\pm16.39$ & $567.32\pm38.76$ \\
CBS       & $3045.78\pm65.48$ & $3506.59\pm110.36$ & $7995.62\pm684.14$ \\
\hline
\end{tabular}
\end{table}
\vspace{-1em}
\subsection{\textit{In Vivo} 3D Human Arms}
We further applied our method to \textit{in vivo} UT arm data (0.25 MHz to 0.6 MHz), collected from two adult volunteers. The experimental setup and instrumentation are shown in Fig.~\ref{fig:fig1}a and explained in the Method section. Arms, with strong scattering effects due to bone, pose a significant challenge for traditional solvers. However, $S^2NO$-FWI preserved accuracy and efficiency: leveraging multiple GPUs (one per slice), it reconstructed an 11-slice (5 cm) arm volume in 8.50 minutes, whereas frequency-domain FWI with conventional solvers took about 60 minutes per slice on the same hardware (Table~\ref{tab:runtime_invivo_main}) (on the same CPU with a single-thread implementation, the runtime is estimated to be approximately 30 hours).

Figure~\ref{fig:fig6}a,b present 3D reconstructions of the arms from both female and male subjects obtained via various methods. The segmented and color‐coded reconstructions demonstrate that the $S^2NO$‐based inversion effectively restored arm anatomy by the distinct delineation of the skin, humerus, blood vessels (e.g., brachial and cephalic veins), and muscles (e.g., biceps brachii and triceps brachii). Moreover, distinct muscle–bone interfaces not only facilitate differentiation between male and female arms based on muscle ratios but also enable the extraction of quantitative characterization of muscle morphology such as cross-sectional area and volume, providing informative metrics for musculoskeletal assessment\cite{stringer2018role,tagliafico2022sarcopenia}. Reference MRI images from the same subjects are provided as a gold-standard comparison. Notably, $S^2NO$ outperformed other methods, delivering high-resolution reconstructions with a parameter-free resolution \cite{descloux2019parameter} of 1.09 mm, closely approaching the MRI standard of 0.88 mm. Sagittal and coronal cross-sectional reconstructions further demonstrate the consistent performance of $S^2NO$ across different imaging planes. In contrast, the numerical solver, CBS, exhibited scattering artifacts near bone structures and distinct black-ring artifacts in arm reconstructions, indicating that traditional numerical solvers are less robust in strongly scattering samples and tend to overfit to observational noise. In comparison, the $S^2NO$-based approach demonstrated superior stability.
\begin{figure}[!htbp]
\centering
\framebox[\columnwidth]{\parbox{\columnwidth}{\includegraphics[width=\columnwidth]{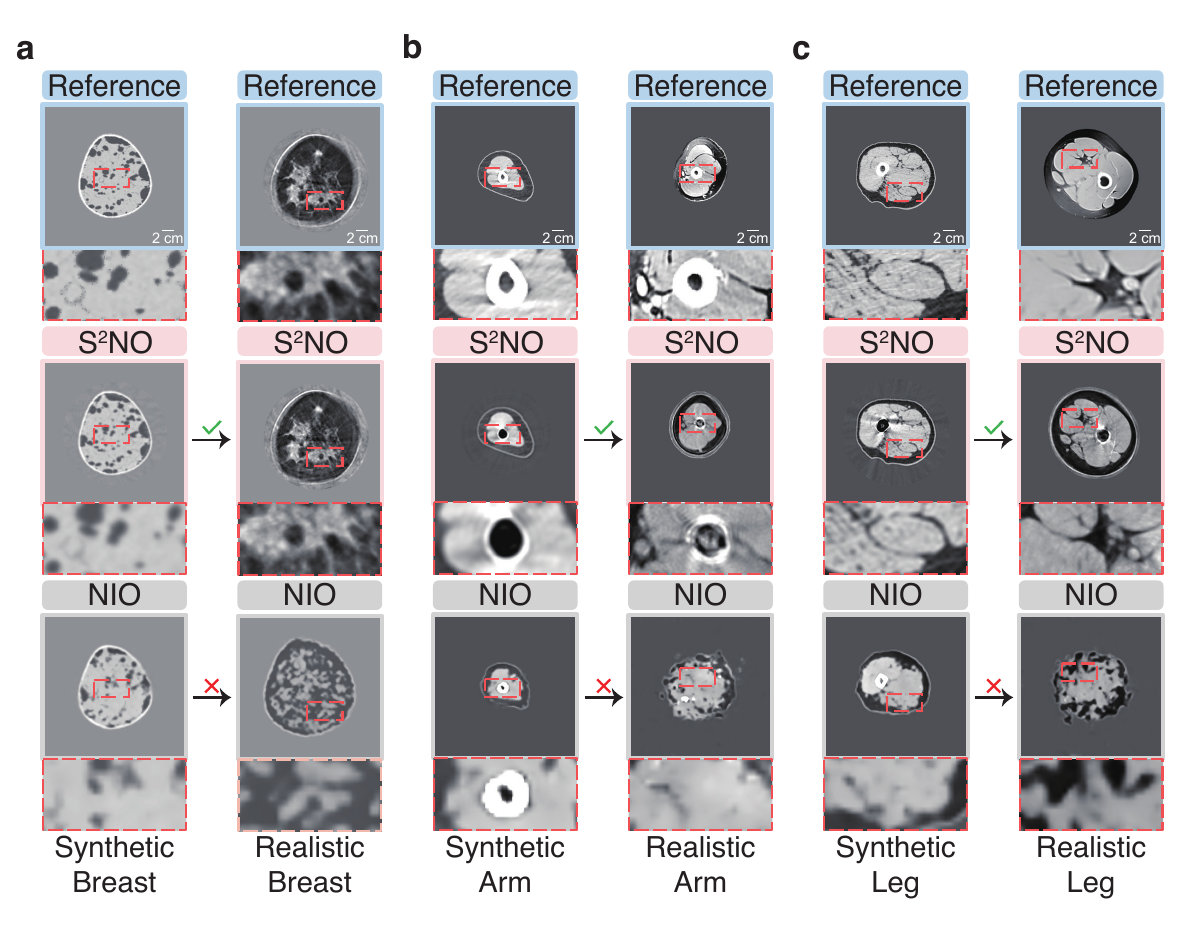}}}
\caption{Comparison of reconstruction quality of synthetic phantoms and \textit{in vivo} tissues between $S^2NO$ and representative direct inversion method, Neural Inverse Operator (NIO)\cite{molinaro2023neural}. The CBS-based reconstruction or the MRI imaging result is provided as reference for \textit{in vivo} data.}
\label{fig:fig8}
\end{figure}
\subsection{\textit{In Vivo} 3D Human Legs}
Finally, we applied our method to high-resolution speed of sound imaging of human legs—a challenging yet crucial task for diagnosing sports injuries (e.g., muscle strain, Achilles tendon rupture) and musculoskeletal conditions (e.g., age-related osteoporosis). Using $S^2NO$-FWI, we achieved high-fidelity 3D reconstructions closely matching reference MRI results (Fig.~\ref{fig:fig6}). For two healthy volunteers, we reconstructed 11 mid-thigh slices spanning approximately 5 cm and combined them to yield a 3D sound speed model. Semantic segmentation of the reconstructed volumes  indicates that $S^2NO$ accurately delineates skin, muscle (e.g., biceps femoris, gracilis, sartorius), fat, blood vessels (e.g., deep femoral vein, great saphenous vein), and femoral structures—including low-density marrow—with a parameter-free resolution of approximately 1.30 mm, an achievement unattainable using classical DAS. The strong scattering in musculoskeletal tissues results in observed signals with extremely low signal-to-noise ratios, complicating the accurate determination of first-arrival times and ultimately causing ToFT to fail in arm and leg imaging applications. Compared to traditional FWI, which required over 133 minutes (around 2.2 hours) using one NVIDIA A800-80GB GPU for reconstruction per slice, $S^2NO$ completed the imaging under 10 minutes. Notably, $S^2NO$-FWI’s speedup over conventional FWI grows from about 6× for arm imaging to 14× for leg imaging (Table~\ref{tab:runtime_invivo_main}), revealing its superior performance on stronger scattering PDEs. This enables successful 3D UT imaging of human legs, underscoring the method’s strong potential for efficient clinical applications where timely diagnosis is critical.

\vspace{-1em}
\section{Discussion}
This article presents a generative neural physics framework that enables rapid, high-resolution \textit{in vivo} UT imaging of breast and musculoskeletal tissues (e.g., arms and legs). By combining generative modeling with physics-informed neural wave simulation, the proposed approach enables accurate sound speed reconstruction in strongly scattering media while significantly reducing computational cost. These properties make UT a practical and accessible modality for quantitative biomechanical imaging.

\subsection{Data- and physics-driven generalization}
Two key innovations drive the success of our framework. First, our cross-modality data generation, combining physics-based modality transfer from CT to ultrasound with diffusion-based augmentation, enables realistic simulation of human organs under experimental UT conditions, even with limited real-world data. This provides a reliable foundation for training and evaluation. Second, our $S^2NO$-based reconstruction addresses the inverse scattering problem by explicitly separating the forward wave physics \(p(y|c)\) from image priors \(p(c)\) within a Bayesian formulation \(p(c|y)\propto p(y|c)p(c)\). Unlike conventional direct inversion networks that implicitly learn \(p(c|y)\), $S^2NO$ robustly models the highly oscillatory wave phenomena inherent in UT, $p(y|c)$ (Fig.~\ref{fig:fig8}) and emphasizes explicit modeling of wave physics rather than merely memorizing structural patterns, resulting in improved generalization to unseen tissues and imaging systems.

\begin{figure}[!htbp]
\centering
\framebox[\columnwidth]{\parbox{\columnwidth}{\includegraphics[width=\columnwidth]{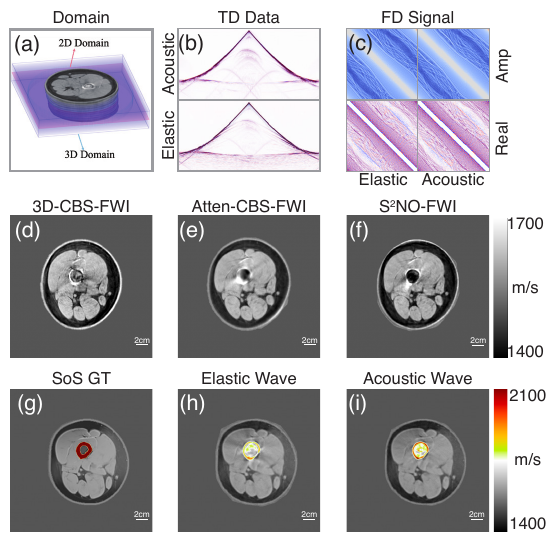}}}
\caption{(a) Schematic illustration of the computational domains used for 2D and 3D FWI. (b) Time-domain signals simulated on the leg phantom constructed from the reconstruction in Fig.~\ref{fig:fig6}, comparing acoustic and elastic wave modelling. (c) Corresponding frequency-domain signals, shown in amplitude and real components. (d)--(f) Reconstructions of the \textit{in vivo} leg data obtained by 3D CBS-FWI, CBS-FWI with joint sound-speed and attenuation reconstruction, and 2D S$^2$NO-FWI, respectively. (g)--(i) Ground-truth sound-speed map of the constructed leg phantom and S$^2$NO-FWI reconstructions from acoustic- and elastic-wave simulated data.}
\label{fig:fig9}
\end{figure}

\subsection{Physics approximation and future directions.}
Despite these advances, the present framework still relies on several simplifying assumptions in wave physics. In particular, attenuation is not explicitly modeled, and wave propagation is approximated within a 2D imaging plane, whereas real tissues support three-dimensional propagation with possible out-of-plane scattering~\cite{lucka2022high}. To assess the influence of these approximations, we performed additional reconstructions on the \textit{in vivo} leg data using a CBS-based forward solver with full 3D modeling and with joint sound-speed-and-attenuation inversion. As shown in Fig.~\ref{fig:fig9}d--f, the resulting reconstructions are broadly consistent with the proposed 2D S$^2$NO-FWI reconstruction in recovering the major musculoskeletal structures. This suggests that, under the annular-array acquisition geometry and the current sub-MHz frequency range, the dominant transmitted longitudinal wavefield is effectively captured by the 2D acoustic approximation. Nevertheless, attenuation and out-of-plane propagation can introduce amplitude, phase, and boundary artifacts, especially near bone interfaces. These results support the practical effectiveness of the current approximation while motivating attenuation-aware and full-3D neural wave solvers for improved quantitative fidelity.

Our current model also considers only acoustic propagation and neglects elastic-wave effects, which can be important in bone-containing tissues~\cite{komatitsch2000wave}. To examine the impact of this approximation, we constructed an MRI-transfer leg phantom and generated both acoustic- and elastic-wave measurements using k-Wave. The simulated time- and frequency-domain signals exhibit visible differences between acoustic and elastic modeling, reflecting the additional wave modes introduced by elasticity (Fig.~\ref{fig:fig9}b,c). However, when these measurements were inverted using the same 2D acoustic S$^2$NO-FWI framework, the recovered sound-speed maps remained similar in their major anatomical structures (Fig.~\ref{fig:fig9}g--i). This observation is consistent with the fact that shear-wave components generated within bone rapidly decay in the surrounding water medium and are therefore less prominent in the measured transmission data. At the same time, elastic effects may still contribute to localized modeling errors around bone boundaries and marrow regions, where acoustic contrast is high and mode conversion is more pronounced.

Finally, the current implementation operates at frequencies lower than those used in conventional clinical reflection-mode ultrasound systems, which typically operate at 3--5 MHz or higher. This choice supports stable transmission-mode sound-speed reconstruction in strongly scattering tissues, but may limit the attainable spatial resolution in some applications. Future work will focus on improving the physical fidelity of the model by incorporating elasticity, attenuation, and full 3D wave propagation, while extending the framework to broader frequency regimes. We also plan to expand the dataset to cover a wider range of organs and tissues, such as brain tissue and skull bone, to enhance robustness in more diverse clinical scenarios.

\ifpeerreview \else
\fi

\clearpage
\input{supplement}

\bibliographystyle{IEEEtran}
\bibliography{references}

\ifpeerreview \else


\begin{IEEEbiographynophoto}{Zhijun Zeng}
received the B.S. degree in mathematics from Shanghai University of Finance and Economics, Shanghai, China. He is currently pursuing the Ph.D. degree with the Yau Mathematical Sciences Center, Tsinghua University, Beijing, China. His research interests include computational imaging, inverse problems, numerical methods for partial differential equations, and machine learning for scientific computing.
\end{IEEEbiographynophoto}

\begin{IEEEbiographynophoto}{Youjia Zheng}
is currently working toward the Ph.D. degree with the College of Future Technology, Peking University, Beijing, China, advised by Prof. He Sun. His research interests include computational wave imaging, physics-guided generative models, and their applications in ultrasound computed tomography.
\end{IEEEbiographynophoto}

\begin{IEEEbiographynophoto}{Chang Su}
received the B.Sc. degree from Tsinghua University, Beijing, China, in 2006, and the M.Sc. and Ph.D. degrees from the Institute of Acoustics, Chinese Academy of Sciences, Beijing, China, in 2009 and 2013, respectively. She is currently a Researcher with the Institute of Acoustics, Chinese Academy of Sciences. Her main research interests include medical ultrasound imaging and simulation of sound fields in complex media.
\end{IEEEbiographynophoto}

\begin{IEEEbiographynophoto}{Qianhang Wu}
received the B.S. degree in engineering from Peking University, Beijing, China, in 2025. He is currently pursuing the master's degree with the School of Mechanics and Engineering Science, Peking University. His research interests include medical ultrasound imaging.
\end{IEEEbiographynophoto}

\begin{IEEEbiographynophoto}{Hao Hu}
is currently working toward the Ph.D. degree with the College of Future Technology, Peking University, Beijing, China, supervised by Prof. He Sun. His research interests include neural operators and applications of artificial intelligence in wave imaging.
\end{IEEEbiographynophoto}

\begin{IEEEbiographynophoto}{Zeyuan Dong}
received the Ph.D. degree in signal and information processing from the University of Chinese Academy of Sciences, Beijing, China, in 2026. His research interests include scientific machine learning, physics-informed neural networks, neural operators, deep learning, partial differential equations, and intelligent computational methods.
\end{IEEEbiographynophoto}

\begin{IEEEbiographynophoto}{Yang Lv}
is currently a Chief Physician with the Department of Orthopedics, Peking University Third Hospital, Beijing, China, and is also affiliated with Peking University Third Hospital Yanqing Hospital, Beijing, China.
\end{IEEEbiographynophoto}

\begin{IEEEbiographynophoto}{Ligang Cui}
is currently a Professor, Chief Physician, and Director of the Department of Ultrasound, Peking University Third Hospital, Beijing, China. His research interests include clinical applications of advanced ultrasound techniques, ultrasound contrast imaging, ultrasound elastography, musculoskeletal ultrasound, and quantitative ultrasound-based assessment of tissue properties.
\end{IEEEbiographynophoto}
\begin{IEEEbiographynophoto}{Zhiyong Hou}
is currently a Professor with the Department of Orthopaedic Surgery, The Third Hospital of Hebei Medical University, Shijiazhuang, China. He also serves as President of The Third Hospital of Hebei Medical University. His clinical and research interests include orthopedic trauma, minimally invasive treatment of pelvic and acetabular fractures, hip fractures, complex limb fractures, and musculoskeletal imaging.
\end{IEEEbiographynophoto}
\begin{IEEEbiographynophoto}{Weijun Lin}
received the B.Sc. degree from Tsinghua University, Beijing, China, in 1994, and the M.Sc. and Ph.D. degrees from the Institute of Acoustics, Chinese Academy of Sciences, Beijing, China, in 1997 and 2003, respectively. He is currently a Researcher with the Institute of Acoustics, Chinese Academy of Sciences. His main research interests include sound fields in complex media, computational acoustics, and medical ultrasound.
\end{IEEEbiographynophoto}
\begin{IEEEbiographynophoto}{Zuoqiang Shi}
received the Ph.D. degree from the Zhou Pei-Yuan Center for Applied Mathematics, Tsinghua University, Beijing, China, in 2008. He is currently a Professor with the Yau Mathematical Sciences Center, Tsinghua University. His main research interests include numerical methods for partial differential equations and their applications, and mathematical image processing.
\end{IEEEbiographynophoto}
\begin{IEEEbiographynophoto}{Yubing Li}
(Member, IEEE) received the B.Sc. degree from Tongji University, Shanghai, China, in 2011, the M.Sc. degree from Universit\'e Paris Diderot--Paris 7, Paris, France, in 2014, and the Ph.D. degree from Paris Sciences \& Lettres (PSL) University, Paris, France, in 2018. He is currently a Researcher with the Institute of Acoustics, Chinese Academy of Sciences, Beijing, China. His main research interests include signal processing, computational imaging, and artificial intelligence research in medical and detection acoustics.
\end{IEEEbiographynophoto}
\begin{IEEEbiographynophoto}{He Sun} (Member, IEEE)
 is an Assistant Professor at the College of Future Technology and the National Biomedical Imaging Center, Peking University, China. Prior to joining Peking University, he was a Postdoctoral Researcher in the Department of Computing and Mathematical Sciences at the California Institute of Technology. He received the Ph.D. degree in Mechanical and Aerospace Engineering from Princeton University in 2019 and the bachelor’s degree in Engineering Mechanics and Economics from Peking University in 2014. His research combines physics-grounded AI algorithms with hardware innovations for imaging at extreme scales. His past work has supported various challenging science missions, such as black hole imaging and the search for Earth-like exoplanets, as well as biomedical and industrial imaging modalities including ultrasound tomography and computational microscopy.
\end{IEEEbiographynophoto}




\fi

\end{document}

%% file: supplement.tex
\clearpage
\onecolumn
\appendices
\setcounter{figure}{0}
\setcounter{table}{0}
\renewcommand{\thefigure}{S\arabic{figure}}
\renewcommand{\thetable}{S\arabic{table}}
\section*{Supplementary Information}

\section*{Conventional UT Reconstruction Algorithm}
\textbf{Delay-and-Sum (DAS)} beamforming is one of the most fundamental and widely used techniques in real-time ultrasound imaging to reconstruct images from raw echo signals received by an array of transducer elements. It aligns and sums the delayed signals from each transducer element to focus on specific points within the imaging plane, thereby reconstructing a 2D structural image of the medium, which represents the echo intensity from tissue scattering. 
Under the full matrix capture (FMC) mode, the transmitting transducer element emits short pulses into the medium, and echoes are received by all transducer elements after bouncing off tissues at various scattering points. For every imaging point, DAS beamforming computes the traveltime from the emitting element to the point and back to each receiver—based on their distances and an assumed constant speed of sound—and applies the corresponding delays before summation. 
Specifically, in DAS imaging we employ the straight-ray (i.e. assuming constant sound speed) and the single-scattering approximations. Under these assumptions, the traveltime from the emitter to a target point and then to the receiver is given by
\begin{equation}
    \tau_{s,r}(\boldsymbol{x}) = \frac{|\boldsymbol{x}_s - \boldsymbol{x}|+|\boldsymbol{x}_r - \boldsymbol{x}|}{c_0},
\end{equation}
where $\boldsymbol{x},\boldsymbol{x}_s$ and $\boldsymbol{x}_r$ denote the coordinates of the imaging point, the source $s$, and the receiver $r$, respectively, and $c_0$ is the (spatially uniform) sound speed. The FMC data are beamformed to reconstruct the spatially varying echo image $I$, reading
\begin{equation}
    I(\boldsymbol{x}) = \sum_{s=1}^{n_s}\sum_{r=1}^{n_r} d_{s,r} (\tau_{s,r}(\boldsymbol{x})),
\end{equation}
where $d_{s,r}$ denotes the recorded signal of time series associated with source $s$ and receiver $r$. Readers are referred to Perrot et al.\cite{perrot2021so} for more technical details.
\textbf{Time of Flight Tomography (ToFT)}
is another powerful imaging technique that leverages ray propagation principles and Time of Flight (ToF) measurements to reconstruct the internal sound speed distribution of imaging targets. In this approach, the physical process is modelled by the Eikonal equation
\begin{equation}
    |\nabla T(\boldsymbol{x})| = \frac{1}{c^2(\boldsymbol{x})},
\end{equation}
where $T$ is the traveltime map from a source location $\boldsymbol{x}_s$ to every location in space subject to the boundary condition of $T(\boldsymbol{x}_s )=0$. Here,$c$ is the spatially varying sound speed. The Eikonal equation can be resolved using fast marching or fast sweeping method.

In this work, we reconstruct the sound speed distribution c using the refraction-corrected ToFT algorithm. The goal is to find an optimal sound speed distribution map $c_{opt} (\boldsymbol{x})$, that best explains traveltimes measured associated with different source-receiver pairs on the ring transducer array. This objective is posed as a least-squares problem:
\begin{equation}
    c_{opt} = \arg\min_c J_{toft} (c) =\frac{1}{2} \sum_{s=1}^{n_s}\sum_{r=1}^{n_r} |t_{s,r}^{cal}(c)-t_{s,r}^{obs}(c)|^2,
\end{equation}
where $t_{s,r}^{cal}$ and $t_{s,r}^{obs}$ are calculated and observed traveltimes, respectively. The gradient of $J_{toft}$ with respect to $c$ is given by
\begin{equation}
    \nabla_c J_{toft} = \sum_{s=1}^{n_s}\sum_{r=1}^{n_r}\frac{\partial t_{s,r}^{cal}}{\partial c} (t_{s,r}^{cal}(c)-t_{s,r}^{obs}(c)),
\end{equation}
where $\frac{\partial t_{s,r}^{cal}}{\partial c}$ is related to the Jacobian matrix, describing how the ToF at a given point changes with respect to variations in the sound speed map $c$. These partial derivatives can be efficiently approximated using ray tracing on the traveltime maps computed via the Eikonal equation. Once the gradient is obtained, the optimization problem is solvable through a gradient based method to estimate the optimal sound speed map $c_{opt}$ that best fits the observed ToFs across different source-receiver pairs.
\section*{ Implementation of Full Waveform Inversion optimization}
We perform full waveform inversion for breast, leg, and arm imaging using a nonlinear conjugate gradient (NCG) optimizer. A hierarchical, multi-scale frequency-marching strategy is leveraged to reconstruct the scattering medium by solving inversion problems sequentially at progressively higher wave frequencies $\omega$. Crucially, each sub-problem leverages the solution from the previous sub-problem as the initialization for the next optimization step. To avoid over-fitting, we apply a gaussian blurring filter to the output of the previous sub-problem before initiating the new one. To ensure the selection of an appropriate initial step size, we normalize both the objective function and its gradient by their initial values within each sub-problem.

Given that the ultrasonic transmitters and receivers in the experimental setup share identical positions, the adjoint states can be directly obtained through a linear combination of the forward simulation results
\begin{equation}
    \lambda_k(\boldsymbol{x}) = [u_i(\boldsymbol{x})]_i [u_i(\boldsymbol{x}_f^{(i)})-y_k^{(i)}]^{\top}_i,
\end{equation}
where $u_i(\boldsymbol{x})$ is the full acoustic wavefield corresponding to the i-th source, $\boldsymbol{x}_f^{(i)}$ defines the i-th source’s location, $y_k^{(i)}\in\mathbb{C}$ is the recorded wavefield data of the i-th transducer. This technique significantly reduces computational costs.

The source intensity and phase of the ultrasound emitted by the instrument are often difficult to measure accurately. To mitigate this uncertainty, we performed source intensity estimation using linear regression during each gradient calculation. Notably, leveraging the linearity of the governing equations with respect to the source term, we used the point source in the forward simulations with unit intensity
\begin{equation}
    \rho_k(\boldsymbol{x}) = \delta(\boldsymbol{x} - \boldsymbol{x}_f^{(k)}).
\end{equation}
The estimated source was obtained by fitting the computed wavefield observation data $u_k(\boldsymbol{x}_f)$ to the measured observation data $y_k$:
\begin{equation}
    \rho_{k}(\boldsymbol{x})=\frac{\sum_{i=1}^{K} y_{k}^{(i)} \overline{u_{k}\left(\boldsymbol{x}_{f}^{(i)}\right)}}{\sum_{i=1}^{K} u_{k}\left(\boldsymbol{x}_{f}^{(i)}\right) \overline{u_{k}\left(\boldsymbol{x}_{f}^{(i)}\right)}} \delta\left(\boldsymbol{x}-\boldsymbol{x}_{f}^{(k)}\right) .
\end{equation}

\section*{Detailed Implementation of $S^2NO$}

\textbf{Encoder and Decoder:} The S$^{2}$NO framework takes two input variables: wave speed \(c\) and input field \(u_{\mathrm{in}}=\rho\).
The input field represents the wavefield generated by the transducer in a water medium for a given frequency and transducer location (i.e., the Helmholtz equation solution with homogeneous media). This input field encodes the frequency and source location of the current problem. We lift the input variables to high-dimensional feature spaces using linear encoders and project the output features of a series of nonlinear layers back to the physical space through MLP decoders. To emulate the Convergent Born Series (CBS)\cite{osnabrugge2016convergent} in the feature space,
we utilize three linear encoders,
\(
q=\mathcal{E}_{\theta_q}(c),
v=\mathcal{E}_{\theta_v}(c),
u_{0}=\mathcal{E}_{\theta_{\mathrm{init}}}(c,\rho),
\)
which introduce latent representations of
\(
q(c)=1-\mathrm{i}\,\nu(c)/\epsilon,
v(c)=\left(\frac{\omega}{c(x)}\right)^{2}-\kappa^{2}-\mathrm{i}\epsilon,
u_{0}(\rho)=\mathcal{G}\rho .
\)

Based on extensive experiments with the network architecture, modifying the dependency of the latent representation \(u_{0}\) to include both \(c\) and \(\rho\) improves performance. This adjustment does not undermine the interpretability or efficiency of $S^2NO$. We will further investigate the underlying reasons for this phenomenon.

\textbf{Spectral Convolution}: Spectral convolution defines a powerful tool for parameterizing complex transformations. Let \(\mathcal{V}\) and \(\mathcal{U}\) denote the input and output function spaces of an operator \(G:\mathcal{V}\to\mathcal{U}\), where any vector-valued function \(v\in\mathcal{V}\) is defined as \(v:\mathcal{D}_{\mathcal{V}}\to\mathbb{R}^{d_{\mathcal{V}}}\) and any vector-valued function \(u\in\mathcal{U}\) is defined as \(u:\mathcal{D}_{\mathcal{U}}\to\mathbb{R}^{d_{\mathcal{U}}}\). The spectral convolution operator \(G\) can be expressed as a linear integral operator:
\begin{equation}
(Gv)(x) := \int_{\mathcal{D}_\mathcal{V}} \kappa(x,y)\, v(y)\, d\nu(y),
\end{equation}
where \(\kappa(x,y)\) is a kernel function. To efficiently parametrize spectral convolution using neural networks, FNO\(^3\) reformulates this operation in the spectral domain and utilizes FFT to implement the integration. Assuming a translation-invariant kernel \(\kappa(x,y)=\kappa(x-y)\) and applying the convolution theorem, the operator becomes:
\begin{equation}
(Gv)(x) = \mathcal{F}^{-1}\!\big(R \cdot \mathcal{F}(v)\big)(x),
\end{equation}
where \(\mathcal{F}\) and \(\mathcal{F}^{-1}\) denote the Fourier and inverse Fourier transforms, respectively. Since we assume that \(\kappa\) is periodic, we parametrize the kernel as a complex-valued tensor \(R\) and directly learn it from the data. For practical implementation, the Fourier series is truncated to a finite number of modes, with the maximum number of modes defined as:
\[
k_{\max}=|Z_{k_{\max}}|
=\Big|\big\{\,k\in\mathbb{Z}^{d}:\ |k_j|\le k_{\max,j},\ \text{for } j=1,\ldots,d \big\}\Big|.
\]
The learned tensor \(R\) has a shape of \((k_{\max}, d_\mathcal{V}, d_\mathcal{U})\) tensor and the spectral multiplication is written as
\[
\big(R\cdot \mathcal{F}(v_\ell)\big)_{k,i}
=\sum_{j=1}^{d_\mathcal{V}} R_{k,i,j}\,\big(\mathcal{F}(v_\ell)\big)_{k,j},\qquad
\forall\, k=1,\ldots,k_{\max},\ i=1,\ldots,d_u .
\]
To ensure sufficient frequency coverage, we analyzed the spectrum of the wavefield and set \(k_{\max}=128\) across all experiments.

\textbf{ $S^2NO$ Layer}

To emulate the Convergent Born Series in the feature space while maintaining the computational efficiency of back-propagation training, we design the $S^2NO$ layer with the following formulation:
\begin{align*}
u_{n+1}
&= BN\!\big(M_{\theta_M}(u_n,q,v)+u_0\big) \\
&= BN\!\left(\Phi_{\theta_M}\!\left((1-q)\,\mathcal{F}^{-1}\!\circ R_{\theta_M}\!\circ \mathcal{F}(v u_n)+q\,u_n\right)+u_0\right),
\end{align*}
where \(BN\) refers to a BatchNorm2D layer, which is employed to accelerate the training of the deep learning model. In the $S^2NO$ model, we utilize the same feature dimension for all layers. The function \(\Phi_{\theta_M}\) is implemented as a two-layer feed-forward network (FFN) with Leaky-ReLU activation. Our experiments revealed that incorporating this non-linearity enhanced the network’s ability to learn complex features.

\textbf{Large scale training}: We trained eight $S^2NO$ models, each tailored to data at a specific frequency. Each model was trained using data from 64 uniformly sampled sources out of a total of 256. All models were optimized with a relative L2 loss function
\begin{equation}
\textit{Relative L2}=\frac{\lVert u-\hat{u}\rVert_{L_2}}{\lVert u\rVert_{L_2}},
\end{equation}
and trained on four NVIDIA A800-80 GB GPUs to ensure fairness. Detailed training parameters are provided in Table~\ref{tab:s1}.

\section*{Implementation Details of Baselines}

All the baselines were trained and tested under the same training strategy: We trained all forward simulation baseline models on the dataset of all three organs (breasts, arms, legs) for 30 epochs using the AdamW optimizer, with an initial learning rate of 5e-3, decayed by a StepLR scheduler. We used relative L2 loss for training and RRMSE for validation. See details in Table~\ref{tab:s1}-\ref{tab:s2}. Next, we will present the implementation of all the baselines.

\textbf{FNO}: To ensure a fair comparison, we used an FNO model with approximately the same number of parameters as the $S^2NO$ model. Specifically, the encoder consists of a two-layer MLP. The Fourier layer employs mode=128 while maintaining a unified width=20. Finally, the decoder maps the latent variables back to the wavefield space using a two-layer MLP.

\textbf{UNet}: We implemented the UNet using the same structure as in the work4 but a larger model size.

\textbf{Neural Inverse Operator}: For inverse imaging, we adopted the Neural Inverse Operator (NIO)\cite{molinaro2023neural} model as a baseline for data‑driven direct inversion and benchmarked it against neural operator–based FWI. NIO combines DeepONet with the FNO to enhance inversion accuracy, and leverages bagging method to improve generalizability. In our implementation, we adjusted the convolutional layer configurations in the branch network to accommodate UT data. Specifically, the DeepONet trunk network comprises an eight‑layer multilayer perceptron (MLP) with 100 hidden neurons per layer to generate 25 basis functions, while the branch network consists of a convolutional neural network (CNN) with 10 Conv2D layers followed by a linear layer to extract the coefficients of these basis functions. The FNO component comprises four Fourier layers with 40 modes and a width of 32.

\textbf{DeepONet}: In this study, we implemented DeepONet for the Helmholtz equation following the architecture of the work\cite{zhang2024blending}: the branch network inputs the tensor $[c(x),Re(u_0 (x)),Im(u_0 (x))]\in \mathbb{R}^{3\times480\times480}$, which is processed by four $3\times3$ Conv2D layers with stride 2 that expand the channel dimension from 3 to 40, 60, 100 and finally 180; the resulting feature maps are then flattened and passed through a three‑layer MLP with hidden sizes of $[180,80,80]$ to produce the coefficients of the basis functions, while the trunk network takes the 2D coordinate $x \in \mathbb{R}^2$ and projects it via a fully connected network with three hidden layers of 80 units each to generate the corresponding basis functions.

\textbf{U-NO}: We implemented U-NO~\cite{rahman2023uno} as an additional neural operator baseline for forward wavefield simulation. U-NO combines a U-shaped multiscale architecture with neural operator layers to capture spatial features at different resolutions. In our implementation, the input and output formats were adapted to the UT forward simulation setting, where the model maps the sound-speed field and source-dependent input field to the complex acoustic wavefield. Specifically, the input is lifted to a base width of 20 by a two-layer MLP and processed by seven spectral operator blocks arranged in a contracting--expanding hierarchy with skip connections, where the channel width is progressively expanded up to $8\times$ the base width and the Fourier modes are set to $[128,64,32,32]$ across scales. A two-layer MLP decoder maps the concatenated latent features back to the real and imaginary parts of the wavefield.

\textbf{BFNO}: We implemented BFNO \cite{zhao2023deep} as a Born-series-inspired neural operator baseline. BFNO follows the iterative structure of the classical Born series and replaces the linear wave-propagation operator with learnable Fourier-domain transformations. This baseline is used to assess the benefit of the CBS-inspired preconditioned architecture adopted in $S^2NO$. In our implementation, the sound-speed map and the source-dependent input field, each appended with grid coordinates, are lifted to a feature width of 20 by linear layers, and seven Born-type iteration layers with 128 Fourier modes update the latent wavefield. A two-layer MLP with 256 hidden units decodes the latent features into the real and imaginary parts of the wavefield.

\textbf{GNOT}: We implemented GNOT~\cite{hao2023gnot} as a Transformer-based neural operator baseline. GNOT employs attention mechanisms to model operator mappings between input and output function spaces. For UT wavefield simulation, the model input was adapted to include the sound-speed map and source-dependent input field, and the output was modified to predict the real and imaginary components of the frequency-domain acoustic wavefield. Our implementation uses two GNOT blocks with a hidden dimension of 64 and 8 attention heads, where each block combines linear cross-attention and self-attention with a mixture of two expert FFNs (MLP ratio 2, GELU activation), and the coordinates and input functions are embedded by separate MLPs.

\textbf{ONO}: We implemented ONO~\cite{xiao2024ono} as another attention-based operator learning baseline. ONO introduces orthogonal attention to improve operator learning and reduce redundancy in the attention representation. We adapted its input and output interfaces to the UT wave simulation task. Our implementation follows the official architecture with 10 attention layers, a hidden dimension of 128, 8 attention heads, and an MLP ratio of 2, where the orthogonal attention uses a 32-dimensional orthonormal basis with Nystr\"om approximation and GELU activation.

\textbf{FFNO}: We implemented FFNO~\cite{tran2023factorized} as a factorized Fourier neural operator baseline. FFNO factorizes multidimensional spectral convolutions into separable one-dimensional Fourier transformations, improving computational efficiency for large-grid operator learning. In our experiments, FFNO was trained under the same data split, optimizer, and loss function as the other forward simulation baselines. Our implementation stacks seven factorized Fourier layers with 128 modes per spatial dimension and a feature width of 20, each wrapped in a residual connection; a two-layer MLP lifts the grid-appended inputs, and a symmetric two-layer MLP projects the latent features to the real and imaginary parts of the wavefield.

\textbf{InversionNet}
: We implemented InversionNet~\cite{wu2019inversionnet} as an end-to-end direct inversion baseline for sound-speed reconstruction. Since the original network operates on time-domain seismic gathers, we adapted it to our frequency-domain UT setting by treating the discrete frequency positions as time positions: the real and imaginary parts of the measurements form two input channels, and the recordings from all sources at all seven frequencies are stacked along the temporal axis, so that all frequencies are processed jointly in a single forward pass. Following the original design, the encoder first compresses this frequency-stacked axis with $7\times1$ and $3\times1$ convolutions of stride 2, then extracts joint spectral--spatial features with $3\times3$ convolutional layers, and finally flattens the feature maps into a 512-dimensional latent vector; all layers are followed by batch normalization and LeakyReLU activation. The decoder mirrors the original deconvolution--convolution design with additional upsampling stages to match our imaging grid, and the final feature map is center-cropped to $480\times480$ and passed through a $3\times3$ convolution to output the sound-speed map.

\textbf{Learned FWI Baseline}: We further compared with the learned full-waveform inversion framework of Lozenski et al.~\cite{lozenski2024learned}, which trains an InversionNet-based encoder--decoder to map USCT measurements directly to speed-of-sound images and augments it with source encoding and a task-informed training objective. In our implementation, the backbone takes the same frequency-domain input arrangement as our InversionNet baseline, treating the discrete frequency positions as time positions and processing the measurements at all seven frequencies jointly. Following the learned-encoder variant of their framework, a source-encoding matrix that linearly superimposes the sequentially excited sources into a smaller set of encoded sources is trained jointly with the reconstruction network, which reduces the input dimensionality and exploits the redundancy of dense UT measurements. The original task-informed objective further augments the image-domain loss with a feature-matching term extracted by a U-Net numerical observer trained for lesion detection; since our datasets provide no diagnostic-task labels for training such an observer, we adopted the purely data-driven form of the training objective, which corresponds to setting the task weight to zero in their formulation and recovers the standard supervised loss.

\subsection*{Evaluation Metrics}
We evaluated the performance of the baseline methods independently for the forward simulation and inverse imaging tasks, using RRMSE for the former and PSNR and SSIM for the latter.

\textbf{Relative RMSE (RRMSE) for forward simulation}: Given the ground-truth physics field $u$ and the model predicted field $\hat{u}$, the RRMSE can be calculated as follows:
\begin{equation}
    RRMSE= \frac{1}{M}\sum_{i=1}^M \frac{\|u_i - \hat{u}_i\|_{L2}}{\|u_i\|_{L2}}.
\end{equation}

\textbf{SSIM for inverse imaging}: Given the ground-truth sound speed $c$ and the reconstructed sound speed $\hat{c}$ , the SSIM of the reconstruction result can be calculated as follows:
\begin{equation}
    SSIM(c,\hat{c}) = \frac{(2\mu_c \mu_{\hat{c}}+C_1)(2\sigma_{c\hat{c}}+C_2)}{(\mu_c^2+\mu_{\hat{c}}^2+C_1)(\sigma_{cc}^2+\sigma_{\hat{c}\hat{c}}^2+C_2)}
\end{equation}
where $\mu_c, \mu_{\hat{c}}$  are means of $c$ and $\hat{c}$, $\sigma_{cc}, \sigma_{\hat{c}\hat{c}}$ are their variances, $\sigma_{c\hat{c}}$ is their covariance, and $C_1,C_2$  are small constants to stabilize the division.

\textbf{Parameter-free Image Resolution Estimation}: To quantitatively assess the effective spatial resolution of the reconstructed images, we employ the parameter-free resolution estimation method based on decorrelation analysis\cite{descloux2019parameter}. The main idea is to identify the highest spatial frequency in a single image that still contains meaningful structural information. Specifically, the reconstructed image is transformed into the Fourier domain, and a sequence of high-pass filtered images with different cutoff frequencies is constructed. The correlation between the original image and each filtered image is then evaluated. As the cutoff frequency increases, structural information is gradually removed and the correlation decays. The cutoff at which this decorrelation indicates the loss of meaningful structure is taken as the effective resolution limit and converted into a spatial resolution value. This method does not require manually selected point targets, edge profiles, or fitting parameters, and thus provides an objective image-based estimate of resolution.

\subsection*{Dataset Filtering and Plausibility Checks}
As described in Sec.~2.1.2, all Stable Diffusion-generated phantoms are filtered before simulation using a set of plausibility checks. These include: \textbf{boundary integrity}, ensuring that the organ mask is connected, complete, closed, and clearly separated from water; \textbf{non-fragmented anatomy}, removing samples with broken tissue regions, isolated components, or disconnected bone/muscle/fat structures; \textbf{topological consistency}, requiring bones to lie inside the limb with plausible number and size, to be surrounded by soft tissue, and for the skin to form the outer layer; and \textbf{physical plausibility}, rejecting samples with unrealistic tissue layouts or severe boundary artifacts. These checks are implemented through rule-based scripts followed by human visual inspection.

\subsection*{Advanced Baseline Comparisons}

To further evaluate the effectiveness of $S^2NO$ for high-frequency wave simulation, we compared it with a broader set of recent neural operator baselines, including U-NO~\cite{rahman2023uno}, BFNO, GNOT~\cite{hao2023gnot}, ONO~\cite{xiao2024ono}, and FFNO~\cite{tran2023factorized}. All models were trained under the same protocol and evaluated at 500 kHz. As shown in Table~\ref{tab:forward_baselines}, $S^2NO$ achieved the lowest RRMSE on both the validation set and the experimental dataset among the compared baseline models under comparable computational cost. Attention-based operator models, including GNOT and ONO, showed limited performance in this setting. This is likely because the UT forward problem considered here involves large spatial grids and highly oscillatory high-frequency wavefields, for which global attention is computationally demanding and difficult to optimize. FFNO improves efficiency through separable factorization of Fourier-domain operations; however, the Green's function kernel of the Helmholtz equation is not generally separable in strongly scattering media, which can introduce an inherent modeling bias. The comparison with BFNO further highlights the importance of the CBS-inspired design: replacing the conventional Born-series structure with the preconditioned CBS formulation improves stability and accuracy in high-frequency, strongly scattering media.

We also evaluated direct inverse imaging baselines on synthetic breast, arm, and leg datasets, including NIO~\cite{molinaro2023neural}, InversionNet~\cite{wu2019inversionnet}, and the learned FWI baseline in~\cite{lozenski2024learned}. As summarized in Table~\ref{tab:inverse_baselines}, these direct inversion models achieved comparable performance across the three anatomical domains. In comparison, $S^2NO$-FWI outperformed all direct inversion baselines, yielding more accurate and structurally consistent reconstructions. These results suggest that explicitly modeling the forward wave physics within an iterative FWI framework provides better generalization than directly learning the measurement-to-image mapping, particularly when the reconstruction involves high-frequency information and strong multiple scattering.

\subsection*{Evaluation of the model performance on Synthetic Dataset}
After training on the synthetic dataset, we evaluated the forward simulation accuracy of our models and baseline models.

\begin{figure*}[!htbp]
\centering
\framebox[\textwidth]{\parbox{0.9\textwidth}{\includegraphics[width=0.9\textwidth]{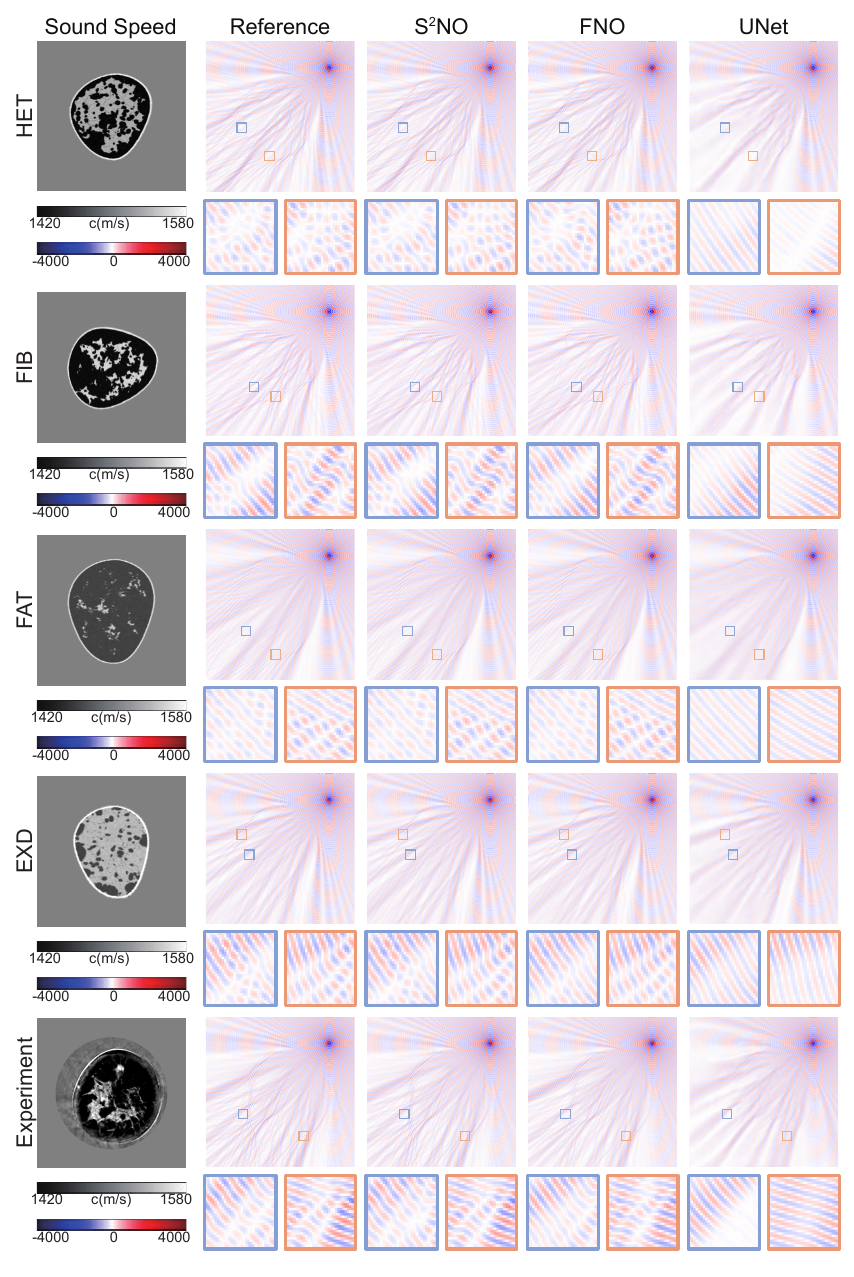}}}
\caption{Forward simulation results for four types of breast phantoms and one \textit{in vivo} human breast.}
\label{fig:s2}
\end{figure*}

\textbf{Breast}: Supplementary Fig.~\ref{fig:s2} displays the wavefield predictions at 0.6 MHz for all models across four distinct breast types, while Supplementary Table~\ref{tab:s6} reports the RRMSE of each model for every category in the test set. Our results demonstrated that $S^2NO$ outperformed all competing baselines at every frequency, accurately capturing the complex scattering phenomena—particularly at the skin–air interface. Furthermore, both quantitative metrics and qualitative assessments confirmed that $S^2NO$ preserved its accuracy when applied to out‑of‑distribution \textit{in vivo} breast tissue, whereas alternative approaches exhibited limited generalizability.

\begin{figure*}[!htbp]
\centering
\framebox[\textwidth]{\parbox{0.9\textwidth}{\includegraphics[width=0.9\textwidth]{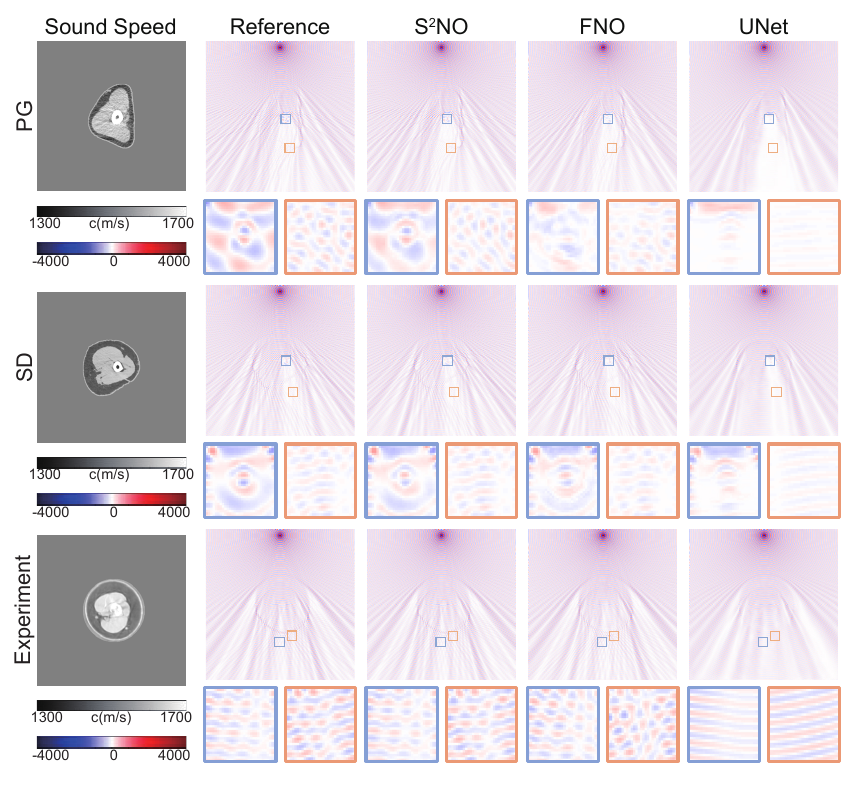}}}
\caption{Forward simulation results for physics-based and Stable Diffusion generated arm phantoms and an \textit{in vivo} human arm.}
\label{fig:s3}
\end{figure*}

\textbf{Arm}: Supplementary Fig.~\ref{fig:s3} presents the 0.6 MHz wavefield predictions by different models for a physics‑based generated arm phantom, a stable diffusion generated arm phantom and an \textit{in vivo} human arm. $S^2NO$ outperformed all baseline methods, exhibiting particularly high accuracy in strongly scattering regions such as bone cavities and the boundary. The RRMSE for each model on these samples, summarized in Supplementary Table~\ref{tab:s7}, further supported this conclusion. Prediction accuracy on the arm dataset was comparatively higher, owing to reduced scattering effects in the smaller arm geometry. Nevertheless, our model retained exceptional generalizability, achieving an RRMSE of 10.2$\%$ at 0.6 MHz. 

\begin{figure*}[!htbp]
\centering
\framebox[\textwidth]{\parbox{0.9\textwidth}{\includegraphics[width=0.9\textwidth]{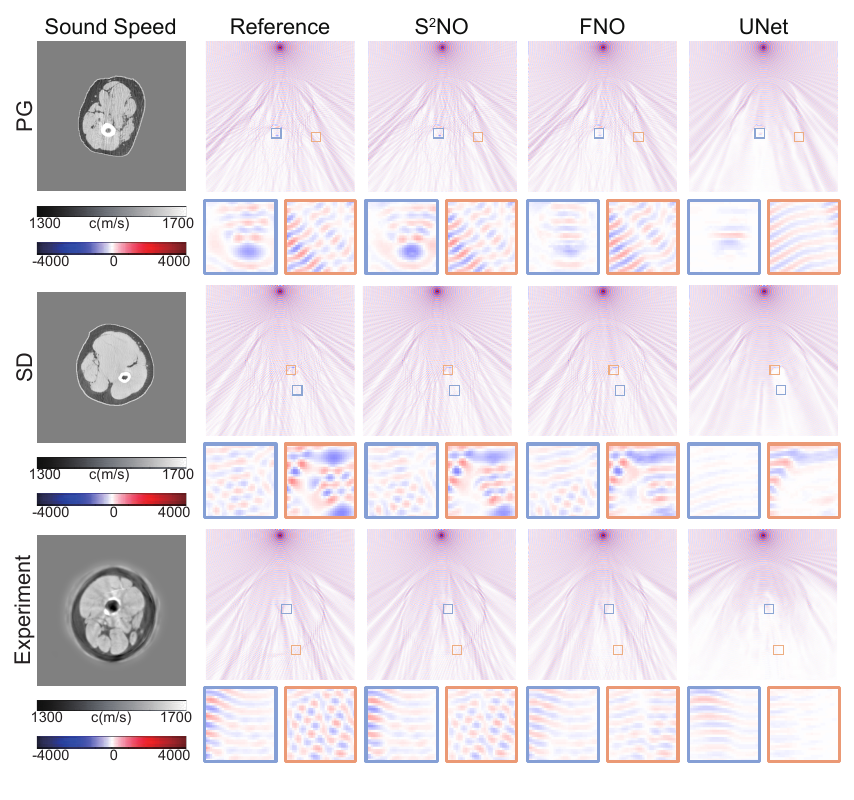}}}
\caption{Forward simulation results for physics-based and Stable Diffusion generated leg phantoms and an \textit{in vivo} human leg.}
\label{fig:s4}
\end{figure*}

\textbf{Leg}: Supplementary Fig.~\ref{fig:s4} presents the 0.6 MHz wavefield predictions provided by different models for a physics‑based generated leg phantom, a stable diffusion–generated leg phantom and an \textit{in vivo} human leg. The RRMSE for each model on these samples is summarized in Supplementary Table~\ref{tab:s8}. Compared to the arm, the leg featured larger scales and more substantial bone structures, resulting in stronger scattering effects. Our $S^2NO$ model demonstrated superior accuracy on both in‑distribution phantoms and out‑of‑distribution clinical sample, accurately capturing scattering phenomena within bone cavities—an aspect vital to the reconstruction quality of musculoskeletal tissues. In contrast, FNO incurred errors of up to $40\%$ at high frequencies, highlighting its inability to effectively exploit high‑frequency information in the reconstruction process.

\subsection*{In Vitro Validation}

\begin{figure*}[!htbp]
\centering
\framebox[\textwidth]{\parbox{0.9\textwidth}{\includegraphics[width=0.9\textwidth]{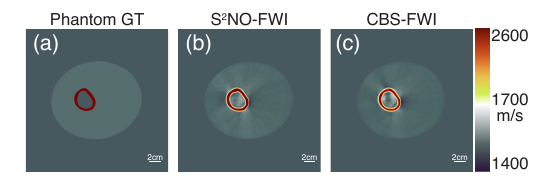}}}
\caption{
(a) Ground-truth sound-speed map of the custom ultrasound phantom. (b) S$^2$NO-FWI reconstruction. (c) CBS-FWI reconstruction. The proposed S$^2$NO-FWI preserves the main sound-speed contrast and produces a reconstruction consistent with CBS-FWI.}
\label{fig:s0}
\end{figure*}

To further assess the quantitative accuracy of the proposed reconstruction framework under controlled experimental conditions, we performed an \textit{in vitro} validation using a custom ultrasound phantom. The data were acquired with the same annular-array system used for the human arm and leg experiments. During acquisition, each transducer element was sequentially excited by a 0.6 MHz bipolar pulse with a peak-to-peak voltage of 180 Vpp. Although the nominal center frequency of the transducer is 900 kHz, we intentionally used a 600 kHz excitation to increase the low-frequency content of the received signals and improve the robustness of full-waveform inversion (FWI). The phantom was positioned at the center of the array and immersed in water. It consists of an inner irregular hollow cylinder and an outer irregular solid cylinder. The inner structure was fabricated from photosensitive resin using stereolithography and was designed to mimic cortical bone, with a sound speed of approximately 2700 m/s. The outer component was made of the same tissue-mimicking acoustic material as that used in the KS107BD ultrasound phantom, with a nominal sound speed of $1540\pm10$ m/s, thereby representing the surrounding soft tissue (see Fig.~\ref{fig:s0}).

Following the initialization strategy used in prior cortical-bone ultrasound tomography studies~\cite{fincke2022quantitative}, we performed frequency-domain FWI with both CBS and S$^2$NO under the same inversion protocol. As shown in Fig.~\ref{fig:s0}, S$^2$NO-FWI accurately recovers the prescribed sound-speed contrast between the resin and tissue-mimicking regions, producing reconstructions that closely agree with those obtained by CBS-FWI. This experiment provides an independent quantitative validation beyond synthetic SSIM evaluation and MRI-based anatomical comparisons, demonstrating that S$^2$NO-FWI preserves physically meaningful sound-speed estimates under controlled out-of-distribution experimental conditions.

\subsection*{Runtime Evaluation}
Supplementary Table~\ref{tab:s10} presents the forward simulation times of CBS, $S^2NO$ and other baseline models at 0.3 MHz and 0.6 MHz. Here, the forward simulation time refers to the computational time required to compute the wavefields for all 256 transducers of a single phantom, with a batch size of 64. All data points were obtained from 100 independent simulations performed on ten randomly selected phantoms.

Supplementary Table~\ref{tab:s11} summarizes the inverse imaging times for synthetic breast, arm, and leg phantoms using CBS, $S^2NO$, and other baseline models. For breast phantoms, we performed a one-round reconstruction using data at seven frequencies (0.3–0.6 MHz). For arm and leg phantoms, we carried out a two-round reconstruction with data at eight frequencies (0.25–0.6 MHz), using the first round’s output—after Gaussian blurring—as the initial model for the second round. All data points were obtained from independent experiments on 5 randomly selected phantoms. For forward simulations, the batch size for all neural‑network models was set to 32. Model loading and compilation times were excluded from the reported durations, as these steps can be completed before the actual imaging process.

Supplementary Table~\ref{tab:s12} summarizes the inverse imaging times for \textit{in vivo} human breast, arm, and leg datasets using CBS, $S^2NO$, and other baseline models. Computation times were measured on two \textit{in vivo} breast datasets, for which reconstructions were performed using data at seven frequencies (0.3–0.6 MHz). Similarly, computation times were evaluated on 11 \textit{in vivo} arm/leg datasets, each reconstructed using data at eight frequencies (0.25–0.6 MHz). The parameters for forward simulation and inference timing were identical to those employed for synthetic data.

\subsection*{Video Visualization}
Slice-by-slice video visualizations of the 3D reconstructions of an \textit{in vivo} volunteer leg and arm are available on our project website: \url{https://open-waves-usct.github.io/}.

\clearpage
\newpage

\begin{figure*}[!htbp]
\centering
\framebox[\textwidth]{\parbox{0.9\textwidth}{\includegraphics[width=0.9\textwidth]{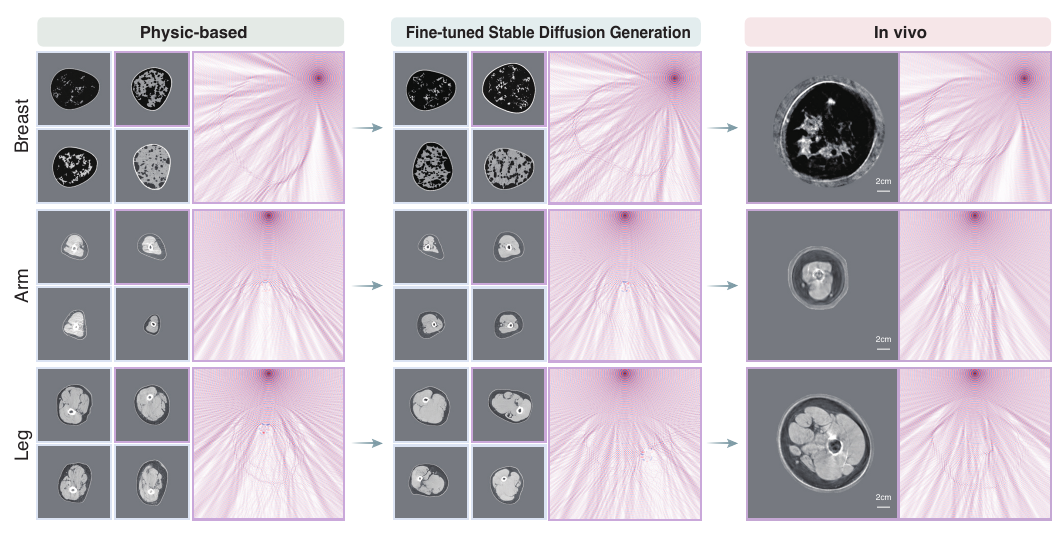}}}
\caption{Representative sound-speed maps of breast, arm, and leg tissues generated by the model and the corresponding wavefields, both closely resembling \textit{in vivo} samples. }
\label{fig:s1}
\end{figure*}

\newpage

\clearpage
\newpage

\clearpage
\newpage

\clearpage
\newpage

\begin{table} 
	\centering
	\caption{\textbf{Hyperparameter list for the implementation of $S^2NO$}}
	\label{tab:s1} 
	\begin{tabular}{@{}l l l@{}}
\hline
\multicolumn{2}{l}{Parameters} & 0.25$\sim$0.6\,MHz \\
\hline
& modes  & 128 \\
& widths & 20  \\
\hline
\multirow{6}{*}{Model Structure}
& layer number & 7 \\
& padding      & 6 \\
& fc\_v        & (3,20,20) \\
& fc\_u0       & (5,20) \\
& fc\_q        & (3,20,20) \\
& fc\_out      & (20,256,2) \\
\hline
\multirow{5}{*}{Training Process}
& optimizer     & AdamW \\
& learning rate & 0.005 \\
& step size     & 6 \\
& gamma         & 0.1 \\
& weight\_decay & 0.00001 \\
\hline
\end{tabular}
\end{table}

\begin{table} 
	\centering
	\caption{\textbf{ Hyperparameter list for training the baseline models}}
	\label{tab:s2} 
\begin{tabular}{@{}l l l@{}}
\hline
Model & Parameters & 0.25$\sim$0.6\,MHz \\
\hline
\multirow{4}{*}{FNO}
& Widths       & 20 \\
& Modes        & 128 \\
& MLP encoder  & (5,20) \\
& MLP decoder  & (3,20) \\
\hline
\multirow{7}{*}{UNet}
& Input channel   & 3 \\
& Output channel  & 2 \\
& Skip channel    & 4 \\
& Depth           & 16 \\
& Channels        & {[}60$\times$4,\;120$\times$4,\;240$\times$4,\;480$\times$4{]} \\
& Activation      & LeakyReLU \\
& Norm            & GroupNorm \\
\hline
\multirow{5}{*}{DeepONet}
& Trunk MLP    & (2,80,80,80) \\
& Activation   & ReLU \\
& BranchNet    & $4$ Conv2D (3$\times$3,channels [3,40,60,100,180]) + MLP(180,80,80) \\
& Basis Number & 80 \\
& Output MLP   & (80,2) \\
\hline
\multirow{6}{*}{NIO}
& Trunk MLP    & (2,100,100,100,100,100,100,100,100,100) \\
& Activation   & GeLU \\
& BranchNet    & $10\times$Conv2D + MLP(100,25) \\
& Basis Number & 25 \\
& Modes        & 40 \\
& Wideth       & 32 \\
\hline
\end{tabular}
\end{table}
\newpage
\begin{table} 
	\centering
	\caption{\textbf{Finetuning/inference parameters of Stable Diffusion}}
	\label{tab:s3} 
\begin{tabular}{@{}l l@{}}
\hline
\multicolumn{1}{c}{\textbf{Finetune Parameter}} & \multicolumn{1}{c}{\textbf{Value}} \\
\hline
Pretrained Model                 & stable-diffusion-v1.4 \\
mixed\_precision                 & fp16 \\
resolution                       & 480 \\
train\_batch\_size               & 1 \\
gradient\_accumulation\_steps    & 3 \\
gradient\_checkpointing          & True \\
learning\_rate                   & 1e-4 \\
snr\_gamma                       & 5.0 \\
lr\_scheduler                    & constant \\
lr\_warmup\_steps                & 0 \\
max\_train\_steps                & 4000 \\
\hline
\multicolumn{1}{c}{\textbf{Inferencing Parameter}} & \multicolumn{1}{c}{\textbf{Value}} \\
\hline
Scheduler & DDIM \\
Step      & 50 \\
Scheduler & DDIM \\
\hline
\end{tabular}
\end{table}
\clearpage
\newpage
\begin{table} 
	\centering
	\caption{\textbf{Summary of breast dataset}}
	\label{tab:s4} 
\begin{tabular}{@{}l l c c c@{}}
\hline
Data Source & Breast Type & \multicolumn{1}{c}{Frequency (MHz)} & Phantoms & Storage \\
\hline
\multirow{4}{*}{VICTRE}
& Heterogeneous (HET)     & \multirow{4}{*}{0.25$\sim$0.6} & 880  & 773GB \\
& Fibroglandular (FIB)    &                                 & 880  & 773GB \\
& Fatty (FAT)             &                                 & 880  & 773GB \\
& Extremely Dense (EXD)   &                                 & 880  & 773GB \\
\hline
\multirow{4}{*}{Stable Diffusion}
& Heterogeneous (HET)     & \multirow{4}{*}{0.25$\sim$0.6} & 1000 & 879GB \\
& Fibroglandular (FIB)    &                                 & 1000 & 879GB \\
& Fatty (FAT)             &                                 & 1000 & 879GB \\
& Extremely Dense (EXD)   &                                 & 1000 & 879GB \\
\hline
\end{tabular}
\end{table}

\begin{table} 
	\centering
	\caption{\textbf{Summary of arm/leg dataset}}
	\label{tab:s5} 
\begin{tabular}{@{}l l c c c@{}}
\hline
Organ & Data Source & \multicolumn{1}{c}{Frequency (MHz)} & Phantoms & Storage \\
\hline
\multirow{2}{*}{Arm}
&  CT Conversion & \multirow{2}{*}{0.25$\sim$0.6} & 809  & 711GB  \\
& Stable Diffusion    &                                   & 6717 & 5.77TB \\
\hline
\multirow{2}{*}{Leg}
&  CT Conversion & \multirow{2}{*}{0.25$\sim$0.6} & 1001 & 880GB  \\
& Stable Diffusion    &                                   & 6000 & 5.15TB \\
\hline
\end{tabular}
\end{table}

\begin{table} 
	\centering
	\caption{\textbf{Forward simulation errors of breast phantoms at 8 frequencies}}
	\label{tab:s6} 
\begin{tabular}{@{}c l c c c c@{}}
\hline
\multirow{2}{*}{Frequency (MHz)} & \multirow{2}{*}{Metric} &
\multicolumn{4}{c}{Model} \\
\cmidrule(lr){3-6}
& & S$^{2}$NO & FNO & UNet & DeepONet \\
\hline
\multirow{2}{*}{0.25} & RRMSE(Val) & \textbf{0.016} & 0.036 & 0.187 & 0.771 \\
                      & RRMSE(Exp) & \textbf{0.034} & 0.089 & 0.258 & 0.891 \\
\hline
\multirow{2}{*}{0.3}  & RRMSE(Val) & \textbf{0.025} & 0.048 & 0.210 & 0.803 \\
                      & RRMSE(Exp) & \textbf{0.049} & 0.114 & 0.295 & 0.925 \\
\hline
\multirow{2}{*}{0.35} & RRMSE(Val) & \textbf{0.033} & 0.061 & 0.234 & 0.822 \\
                      & RRMSE(Exp) & \textbf{0.067} & 0.145 & 0.326 & 0.954 \\
\hline
\multirow{2}{*}{0.4}  & RRMSE(Val) & \textbf{0.043} & 0.072 & 0.244 & 0.835 \\
                      & RRMSE(Exp) & \textbf{0.085} & 0.157 & 0.348 & 0.931 \\
\hline
\multirow{2}{*}{0.45} & RRMSE(Val) & \textbf{0.049} & 0.087 & 0.269 & 0.844 \\
                      & RRMSE(Exp) & \textbf{0.094} & 0.189 & 0.405 & 0.928 \\
\hline
\multirow{2}{*}{0.5}  & RRMSE(Val) & \textbf{0.057} & 0.101 & 0.281 & 0.851 \\
                      & RRMSE(Exp) & \textbf{0.104} & 0.206 & 0.406 & 0.935 \\
\hline
\multirow{2}{*}{0.55} & RRMSE(Val) & \textbf{0.067} & 0.113 & 0.302 & 0.856 \\
                      & RRMSE(Exp) & \textbf{0.131} & 0.222 & 0.444 & 0.931 \\
\hline
\multirow{2}{*}{0.6}  & RRMSE(Val) & \textbf{0.077} & 0.134 & 0.310 & 0.862 \\
                      & RRMSE(Exp) & \textbf{0.139} & 0.263 & 0.454 & 0.926 \\
\hline
\end{tabular}
\end{table}
\clearpage
\newpage
\begin{table} 
	\centering
	\caption{\textbf{Forward simulation errors of arm phantoms at 8 frequencies}}
	\label{tab:s7} 
\begin{tabular}{@{}c l c c c c@{}}
\hline
\multirow{2}{*}{Frequency (MHz)} & \multirow{2}{*}{Metric} &
\multicolumn{4}{c}{Model} \\
\cmidrule(lr){3-6}
& & S$^{2}$NO & FNO & UNet & DeepONet \\
\hline
\multirow{2}{*}{0.25} & RRMSE(Val) & \textbf{0.029} & 0.051 & 0.183 & 0.531 \\
                      & RRMSE(Exp) & \textbf{0.015} & 0.038 & 0.217 & 0.574 \\
\hline
\multirow{2}{*}{0.3}  & RRMSE(Val) & \textbf{0.040} & 0.068 & 0.201 & 0.540 \\
                      & RRMSE(Exp) & \textbf{0.021} & 0.052 & 0.465 & 0.580 \\
\hline
\multirow{2}{*}{0.35} & RRMSE(Val) & \textbf{0.052} & 0.079 & 0.218 & 0.545 \\
                      & RRMSE(Exp) & \textbf{0.028} & 0.084 & 0.338 & 0.521 \\
\hline
\multirow{2}{*}{0.4}  & RRMSE(Val) & \textbf{0.058} & 0.087 & 0.222 & 0.546 \\
                      & RRMSE(Exp) & \textbf{0.039} & 0.109 & 0.294 & 0.648 \\
\hline
\multirow{2}{*}{0.45} & RRMSE(Val) & \textbf{0.066} & 0.100 & 0.236 & 0.548 \\
                      & RRMSE(Exp) & \textbf{0.053} & 0.130 & 0.268 & 0.653 \\
\hline
\multirow{2}{*}{0.5}  & RRMSE(Val) & \textbf{0.075} & 0.108 & 0.243 & 0.552 \\
                      & RRMSE(Exp) & \textbf{0.063} & 0.132 & 0.342 & 0.657 \\
\hline
\multirow{2}{*}{0.55} & RRMSE(Val) & \textbf{0.071} & 0.113 & 0.255 & 0.556 \\
                      & RRMSE(Exp) & \textbf{0.068} & 0.142 & 0.393 & 0.668 \\
\hline
\multirow{2}{*}{0.6}  & RRMSE(Val) & \textbf{0.077} & 0.123 & 0.260 & 0.560 \\
                      & RRMSE(Exp) & \textbf{0.102} & 0.186 & 0.348 & 0.676 \\
\hline
\end{tabular}
\end{table}
\begin{table} 
	\centering
	\caption{\textbf{Forward simulation errors of leg phantoms at 8 frequencies}}
	\label{tab:s8} 
\begin{tabular}{@{}c l c c c c@{}}
\hline
\multirow{2}{*}{Frequency (MHz)} & \multirow{2}{*}{Metric} &
\multicolumn{4}{c}{Model} \\
\cmidrule(lr){3-6}
& & S$^{2}$NO & FNO & UNet & DeepONet \\
\hline
\multirow{2}{*}{0.25} & RRMSE(Val) & \textbf{0.042} & 0.078 & 0.248 & 0.733 \\
                      & RRMSE(Exp) & \textbf{0.054} & 0.164 & 0.773 & 0.878 \\
\hline
\multirow{2}{*}{0.3}  & RRMSE(Val) & \textbf{0.062} & 0.104 & 0.271 & 0.737 \\
                      & RRMSE(Exp) & \textbf{0.076} & 0.205 & 0.527 & 0.878 \\
\hline
\multirow{2}{*}{0.35} & RRMSE(Val) & \textbf{0.080} & 0.122 & 0.293 & 0.744 \\
                      & RRMSE(Exp) & \textbf{0.096} & 0.226 & 0.734 & 0.899 \\
\hline
\multirow{2}{*}{0.4}  & RRMSE(Val) & \textbf{0.099} & 0.139 & 0.304 & 0.757 \\
                      & RRMSE(Exp) & \textbf{0.170} & 0.269 & 0.606 & 0.970 \\
\hline
\multirow{2}{*}{0.45} & RRMSE(Val) & \textbf{0.106} & 0.161 & 0.326 & 0.773 \\
                      & RRMSE(Exp) & \textbf{0.138} & 0.300 & 0.507 & 0.915 \\
\hline
\multirow{2}{*}{0.5}  & RRMSE(Val) & \textbf{0.118} & 0.181 & 0.336 & 0.760 \\
                      & RRMSE(Exp) & \textbf{0.146} & 0.316 & 0.398 & 0.905 \\
\hline
\multirow{2}{*}{0.55} & RRMSE(Val) & \textbf{0.127} & 0.195 & 0.358 & 0.764 \\
                      & RRMSE(Exp) & \textbf{0.183} & 0.342 & 0.512 & 0.904 \\
\hline
\multirow{2}{*}{0.6}  & RRMSE(Val) & \textbf{0.139} & 0.213 & 0.361 & 0.767 \\
                      & RRMSE(Exp) & \textbf{0.187} & 0.419 & 0.547 & 0.916 \\
\hline
\end{tabular}
\end{table}

\begin{table}
    \centering
    \caption{\textbf{Forward baseline comparison and architectural ablation at 500 kHz}}
    \label{tab:forward_baselines}
\begin{tabular}{@{}l c c c c c c c c c c@{}}
\hline
\textbf{Metric}
& \textbf{S$^2$NO}
& \textbf{U-NO}
& \textbf{BFNO}
& \textbf{GNOT}
& \textbf{ONO}
& \textbf{FFNO}
& \begin{tabular}[c]{@{}c@{}}\textbf{Shared}\\\textbf{Param.}\end{tabular}
& \begin{tabular}[c]{@{}c@{}}\textbf{Width}\\\textbf{20$\to$40}\end{tabular}
& \begin{tabular}[c]{@{}c@{}}\textbf{Layer}\\\textbf{7$\to$5}\end{tabular}
& \begin{tabular}[c]{@{}c@{}}\textbf{Single}\\\textbf{Organ}\end{tabular}
\\
\hline
RRMSE (Val) $\downarrow$
& \textbf{0.083} & 0.185 & 0.112 & 0.607 & 0.356 & 0.211 & 0.111 & 0.068 & 0.121 & 0.090 \\
RRMSE (Exp) $\downarrow$
& \textbf{0.104} & 0.252 & 0.156 & 0.822 & 0.636 & 0.300 & 0.140 & 0.106 & 0.172 & 0.178 \\
Speed
& $1\times$ & $1.29\times$ & $0.95\times$ & $2.61\times$ & $1.62\times$ & $1.51\times$ & $1\times$ & $1.92\times$ & $0.78\times$ & $1\times$ \\
\hline
\end{tabular}
\end{table}

\begin{table}
    \centering
    \caption{\textbf{Inverse baseline comparison on synthetic datasets}}
    \label{tab:inverse_baselines}
\begin{tabular}{@{}l c c c c@{}}
\hline
\textbf{Method}
& \textbf{NIO}
& \textbf{InversionNet}
& \textbf{\cite{lozenski2024learned}}
& \textbf{S$^2$NO}
\\
\hline
SSIM Breast / Arm / Limb $\uparrow$
& 0.87 / 0.96 / 0.90
& 0.85 / 0.94 / 0.88
& 0.88 / 0.94 / 0.90
& \textbf{0.96 / 0.98 / 0.97} \\
\hline
\end{tabular}
\end{table}

\clearpage
\newpage
\begin{table} 
    \centering
    \caption{\textbf{Reconstruction quality of breast/arm/leg phantoms using different models}}
    \label{tab:s9}
\begin{tabular}{@{}l l c c c c@{}}
\hline
\multirow{2}{*}{Organ} & \multirow{2}{*}{Metric} & \multicolumn{4}{c}{Model} \\
\cline{3-6}
& & CBS & S$^{2}$NO & FNO & UNet \\
\hline
Breast & SSIM$\uparrow$ & $0.951\pm0.052$ & \textbf{$0.958\pm0.031$} & $0.938\pm0.034$ & $0.921\pm0.029$ \\
Arm    & SSIM$\uparrow$ & \textbf{$0.980\pm0.006$} & $0.978\pm0.007$ & $0.971\pm0.008$ & $0.974\pm0.009$ \\
Leg    & SSIM$\uparrow$ & \textbf{$0.969\pm0.005$} & $0.966\pm0.006$ & $0.958\pm0.013$ & $0.892\pm0.038$ \\
\hline
\end{tabular}
\end{table}

\begin{table}
    \centering
    \caption{\textbf{Evaluation of runtime for forward simulation}}
    \label{tab:s10}
\begin{tabular}{@{}l c c c c c@{}}
\hline
\multirow{2}{*}{Model} & \multirow{2}{*}{Frequency} & \multirow{2}{*}{\#Parameter} & \multicolumn{3}{c}{Inference time [s]} \\
\cline{4-6}
 &  &  & Breast & Arm & Leg \\
\hline
S$^{2}$NO & All & 367M &  & $1.3052\pm0.0002$ &  \\
FNO       & All & 367M &  & $0.6102\pm0.0002$ &  \\
UNet      & All & 25M  &  & $0.8404\pm0.0004$ &  \\
DeepONet  & All & 5M   &  & $0.02290\pm0.00001$ &  \\
\hline
CBS & 0.3\,MHz & / & $7.619\pm0.430$ & $19.241\pm2.610$ & $21.773\pm1.183$ \\
CBS & 0.6\,MHz & / & $13.718\pm2.847$ & $31.894\pm4.891$ & $44.734\pm7.443$ \\
\hline
\end{tabular}
\end{table}

\begin{table}
    \centering
    \caption{\textbf{Evaluation of runtime for inverse imaging process of synthetic data}}
    \label{tab:s11}
\begin{tabular}{@{}l c c c c@{}}
\hline
\multicolumn{5}{c}{Inverse Imaging of synthetic data} \\
\hline
\multirow{2}{*}{Model} & \multirow{2}{*}{\# Parameters} & \multicolumn{3}{c}{Imaging process time [s]} \\
\cline{3-5}
& & Breast & Arm & Leg \\
\hline
S$^{2}$NO & 367M & $176.56\pm25.12$ & $393.29\pm8.08$  & $482.87\pm13.73$ \\
FNO       & 367M & $109.84\pm11.60$ & $260.15\pm19.78$ & $290.26\pm17.18$ \\
UNet      & 25M  & $149.00\pm20.33$ & $281.09\pm3.77$  & $334.45\pm93.93$ \\
CBS       & /    & $963.04\pm37.35$ & $5807.95\pm160.85$ & $6308.23\pm496.15$ \\
\hline
\end{tabular}
\end{table}

\begin{table}
    \centering
    \caption{\textbf{Evaluation of runtime for inverse imaging process of experimental data}}
    \label{tab:s12}
\begin{tabular}{@{}l c c c c@{}}
\hline
\multicolumn{5}{c}{Inverse Imaging of experimental data} \\
\hline
\multirow{2}{*}{Model} & \multirow{2}{*}{\# Parameters} & \multicolumn{3}{c}{Imaging process time [s]} \\
\cline{3-5}
& & Breast & Arm & Leg \\
\hline
S$^{2}$NO & 367M & $466.85\pm74.76$ & $510.29\pm16.39$ & $567.32\pm38.76$ \\
CBS       & /    & $3045.78\pm65.48$ & $3506.59\pm110.36$ & $7995.62\pm684.14$ \\
\hline
\end{tabular}
\end{table}
\clearpage
\newpage